\documentclass[times,twocolumn,final,authoryear]{elsarticle}
\makeatletter\if@twocolumn\PassOptionsToPackage{switch}{lineno}\else\fi\makeatother

\makeatletter
\usepackage{prletters}
\long\def\MaketitleBox{%
  \resetTitleCounters
  \def\baselinestretch{1}%
  \begin{\elsarticletitlealign}%
   \def\baselinestretch{1}%
    \vspace*{7.5pc}
    \Large\@title\par\vskip18pt
    \normalsize
  \ifdoubleblind
    \vspace*{2pc}
  \else
    \elsauthors\par\vskip10pt
    {\footnotesize\itshape\elsaddress}\par\vskip12pt
  \fi
\vspace*{-\baselineskip}%
\rule{.8\textwidth}{.2pt}\vskip10pt
    \begin{tabular*}{.8\textwidth}{c@{}p{.9\textwidth}}
     \hspace*{7pt}&ABSTRACT\\[8pt]
    \hline\\[-8pt]
    \end{tabular*}
    \hspace*{13.5pt}\parbox[t]{.777\textwidth}{\unhbox\absbox}
  \end{\elsarticletitlealign}%
\rule{.8\textwidth}{.2pt}
  \vspace*{2pc}
}
\makeatother

\usepackage{framed,multirow}
\definecolor{newcolor}{rgb}{.8,.349,.1}
\newif\ifdoubleblind\doubleblindfalse
\journal{Pattern Recognition Letters}
\def\correspAuthor#1{\gdef\corresponding{#1}}
\correspAuthor{}

\usepackage{tabulary,xcolor}
\usepackage{amsfonts,amsmath,amssymb}
\usepackage[T1]{fontenc}
\makeatletter
\let\save@ps@pprintTitle\ps@pprintTitle
\def\ps@pprintTitle{\save@ps@pprintTitle\gdef\@oddfoot{\footnotesize\itshape \null\hfill\today}}
\def\hlinewd#1{%
  \noalign{\ifnum0=`}\fi\hrule \@height #1%
  \futurelet\reserved@a\@xhline}

\AtBeginDocument{\ifNAT@numbers \biboptions{sort&compress}\fi}
\makeatother

\usepackage{graphicx}
\usepackage{appendix}
\usepackage{subfigure}
\usepackage{amsmath}
\usepackage{mathptmx} 
\usepackage{lipsum}
\usepackage[hyphens]{url}
\usepackage[hidelinks]{hyperref}
\usepackage{flushend}

\begin{document}

\begin{frontmatter}

\title{
    Detecting Owner-member Relationship with Graph Convolution Network in Fisheye Camera System    
}
    
\author[a]{Zizhang Wu}
\ead{zizhang.wu@zongmutech.com}
\author[a]{Jason Wang}
\author[b]{Tianhao Xu}
\ead{tianhao.xu@tu-braunschweig.de}
\author[a]{Fan Wang}
    
\address[a]{
    Zongmu Technology\unskip, Shanghai\unskip, China}
  	
\address[b]{
    Braunschweig University of Technology\unskip, Braunschweig\unskip, Germany}

\begin{abstract}
 The owner-member relationship between wheels and vehicles contributes significantly to the 3D perception of vehicles, especially in embedded environments. However, to leverage this relationship we must face two major challenges: i) Traditional IoU-based heuristics have difficulty handling occluded traffic congestion scenarios. ii) The effectiveness and applicability of the solution in a vehicle-mounted system is difficult. To address these issues, we propose an innovative relationship prediction method, \textbf{DeepWORD}, by designing a graph convolutional network (GCN). Specifically, to improve the information richness, we use feature maps with local correlation as input to the nodes. Subsequently, we introduce a graph attention network (GAT) to dynamically correct the a priori estimation bias. Finally, we designed a dataset as a large-scale benchmark which has annotated owner-member relationship, called \textbf{WORD}. In the experiments we learned that the proposed method achieved state-of-the-art accuracy and real-time performance. The \textbf{WORD} dataset is made publicly available at \url{https://github.com/NamespaceMain/ownermember-relationship-dataset.}\vspace{10pt}
~
\end{abstract}

\end{frontmatter}
    
\section{INTRODUCTION}
\begin{figure}[!h]
    \centering
    \includegraphics[width=7.5cm]{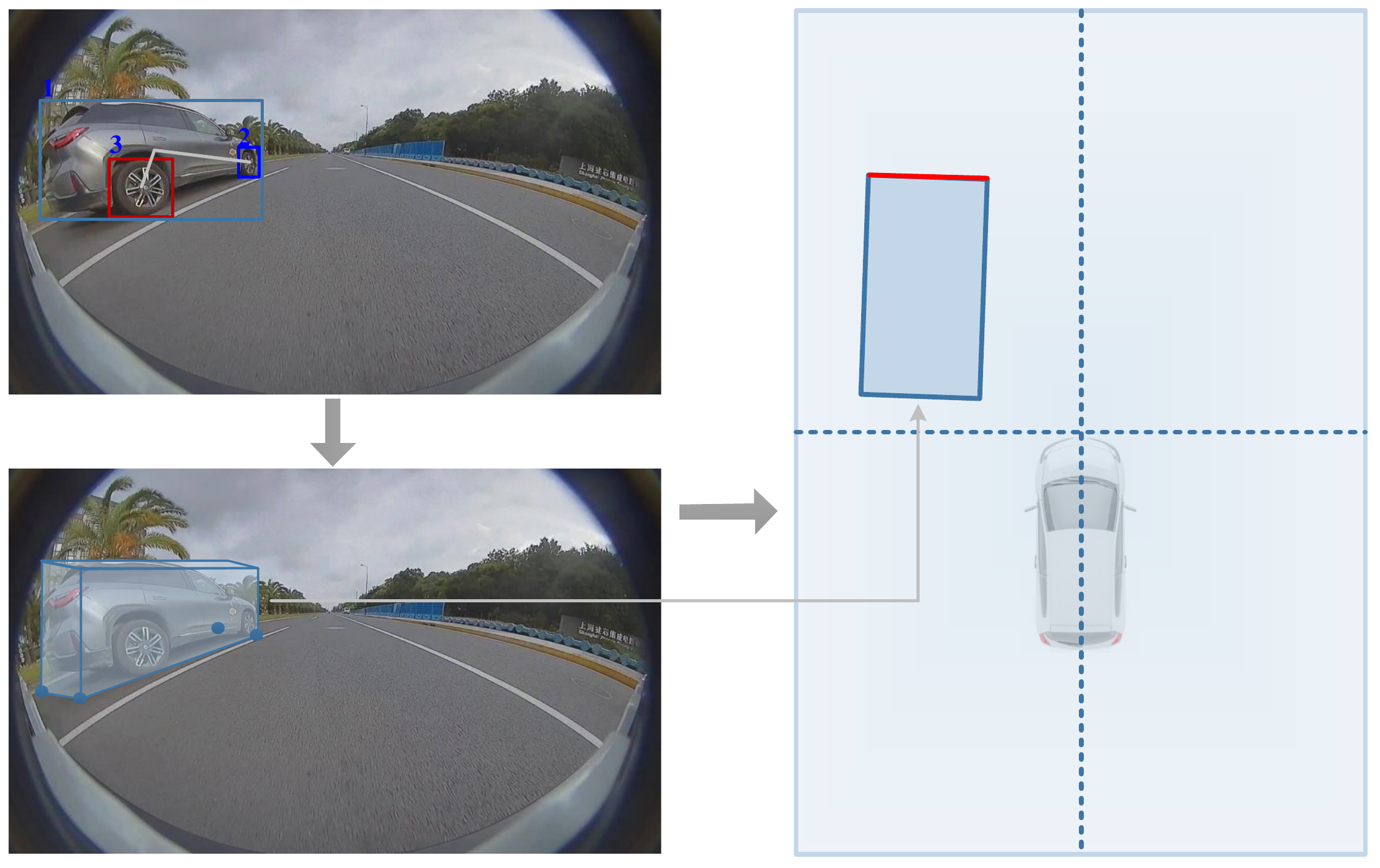}
    \caption{The visualization of the owner-member relationship between wheels and vehicles on 3D information acquisition. The estimation of four BEV corners come into being after obtaining the relationship between wheels and vehicles. In this way, the projection position of surrounding vehicles is helpful for the autonomous driving system to make decision.}
    \label{fig:Figure01}
\end{figure}

\begin{figure}[!t]
    \centering
    \includegraphics[width=8cm]{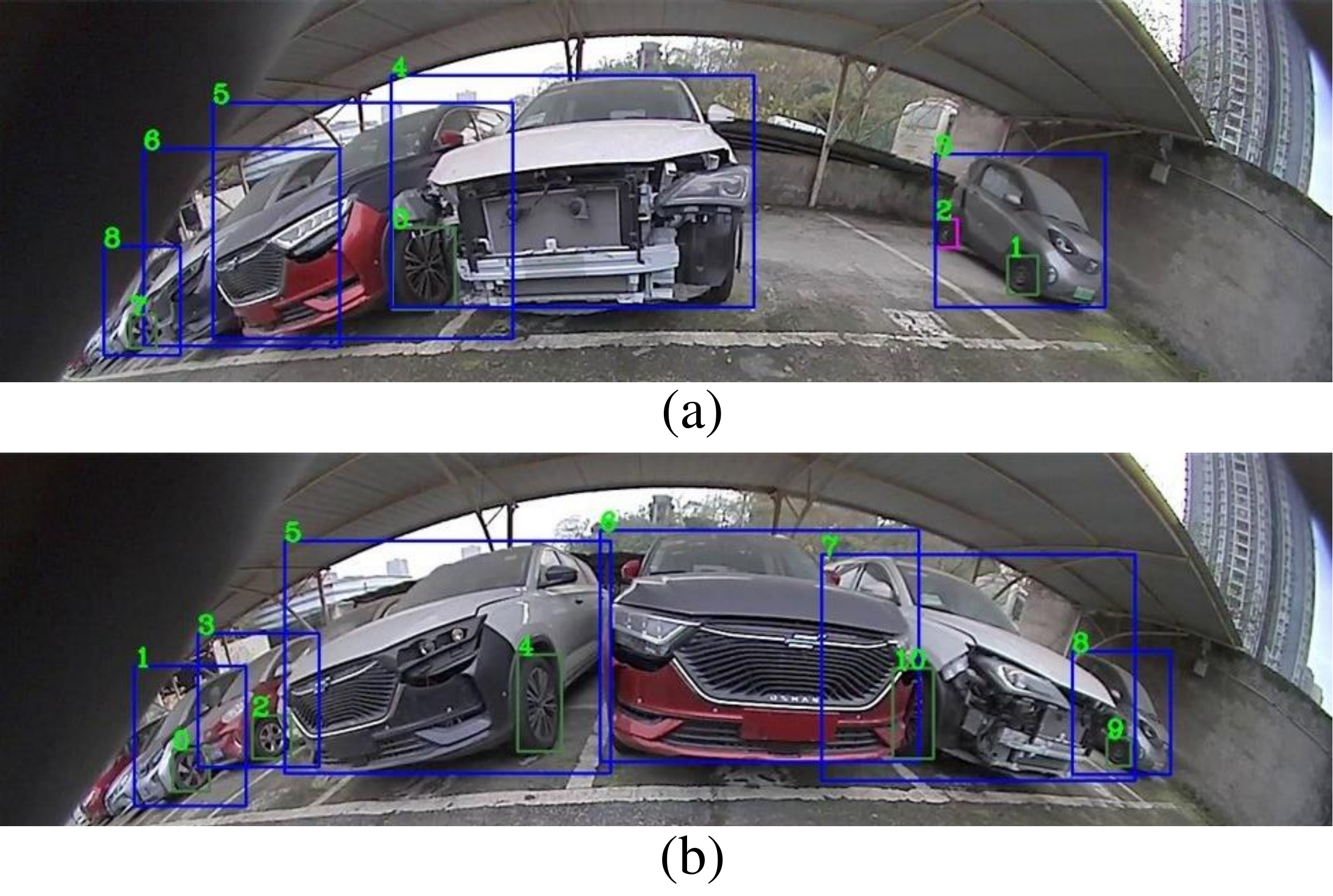}
    \caption{The visualization of complex scenarios. It is obvious that the bounding-boxes of vehicles 4 and 5 completely contain wheel 0 as shown in (a), and the bounding-boxes of vehicles 7 and 8 also completely contain wheel 9 as shown in (b). In these scenarios, the prediction of the owner-member relationship is not available with the IoU alone.}
    \label{fig:Figure02}
\end{figure}

Autonomous vehicles \cite{article13} are rapidly evolving, and various aspects of their applications are attracting increasing attention, such as 3D perception in vehicle-mounted systems \cite{article14, article15}, which is critical for localisation \cite{article16, article17}, planning, obstacle avoidance \cite{article18, article19}, etc., as shown in Figure \ref{fig:Figure01}. The method \cite{article20, article21, article22} using the homography transformation \cite{article20} of wheel grounding points is currently dominant. They estimate the localization of the target vehicle by detecting the wheels in a limited low-cost environment. In complex scenarios where different vehicles overlap each other, the use of wheel and vehicle owner-member relationships can effectively counteract this confusion. 
IoU is a method commonly used to predict relationships between different objects, but the accuracy of this method is very low for complex scenarios, as shown in Figure \ref{fig:Figure02}, in which one wheel belongs to two different vehicle bounding boxes at the same time.

In this study, to address the problem of vehicle overlap in complex scenarios and to constraint the shortcomings of prior analyses, we propose a model with a graph convolutional network(GCN) structure to predict the owner-member relationship, which can implicitly learn the relationship between the wheels and vehicles.

\section{RELATED WORK AND OUR CONTRIBUTIONS}
\sloppy{}

\subsection{Related work}
\paragraph{Traditional relationship prediction methods}

Models designed using traditional methods usually rely on statistical and design thresholds through data such as semantics, pose, and order constraints of objects and predict the relationship between objects. \cite{article3} employed information about the spatial layout of detected objects to predict the relationship between objects. To predict the correlation between objects,  \cite{article4} uses a priori statistics of the owner-member relationship between objects obtained through Conditional Random Field (CRF) \cite{article25}. \cite{article5} used a priori knowledge to train the model to learn objects that frequently occur in pairs, and LSTM (Long short-term memory (LSTM) \cite{article26} to encode the images to enhance contextual information and thus improve the performance of relationship prediction.

\paragraph{Graph convolutional network methods}

By further exploration of the method \cite{article6, article7, article8, article9}, GCN can be used in a frequency domain based mathematical representation paradigm, so that graph convolution has the properties of general convolutional structures can be used in deep learning training. Owner-member and relative location relationship prediction is the main target of GCN relationship prediction. \cite{article10} combines the understanding of aggregation from a spatial domain perspective into the training of graph neural networks to update the graph convolution. This approach is called graph SAGE. In addition, it generalizes the expressions in the spatial domain and optimizes the difficulties encountered in the training of large-scale graph data institutions. \cite{article11} added attention mechanism in graph neural network to improve the interaction between nodes. \cite{article12} introduces a spatial-aware graph relationship network (SGRN) to encode graph relationships to automatically identify relationships between targets. It combines important semantic information and object space relations to get better relationship predictions

\subsection{Our motivations and contributions}

After analysing the above methods, there is still great potential in how to improve the prediction accuracy of the owner-member relationship between the wheel and the vehicle.

The first challenge is that autonomous driving does not yet have a available public dataset in the field of relationship prediction, which is also an impossible task for individuals. Secondly, the application of a priori statistics to predict the owner-membership relationships of wheels and vehicles, as mentioned in the above approach, is only applicable to single scenario and cannot solve complex real-world scenarios. Therefore, it is also difficult to produce positive effects in practical autonomous driving. Moreover, the stacked logical judgments will reduce the computational efficiency of the system, thus increasing the safety risks.

In this study, we draw on previous methods and make corresponding improvements and optimizations to enhance the prediction accuracy. Specifically, we propose a GCN-based owner-membership relationship prediction method for wheels and vehicles, which models the owner-membership relationship between wheels and vehicles by regarding the geometric relative positions between wheels and vehicles as a priori statistics with the GCN structure. In addition, to improve the information richness of the nodes in the GCN, we use feature vectors of local correlations as the input to the nodes. Finally, we introduce the graph attention network module (GAT) \cite{article11} to counteract the effect of noise on accuracy, because GAT can dynamically amend the deviation of prior estimates of edges through the training process. Our contributions are summarised as follows:

1) To attract the participation of more researchers in the field, we have created a large-scale benchmark dataset WORD containing 9000 samples.

2) We propose a GCN-based wheel and vehicle owner-member relationship prediction network that is more applicable to relationship prediction in complex scenarios.

3) We validate the effectiveness of our proposed method using the WORD dataset. The experimental results show that this method achieves superior accuracy, especially real-time effects, in an embedded environment.

\section{DEEPWORD: A GCN-BASED APPROACH FOR OWNER-MEMBER RELATIONSHIP DETECTION}

\begin{figure*}[!t]
    \centering
    \includegraphics[width=0.95\textwidth]{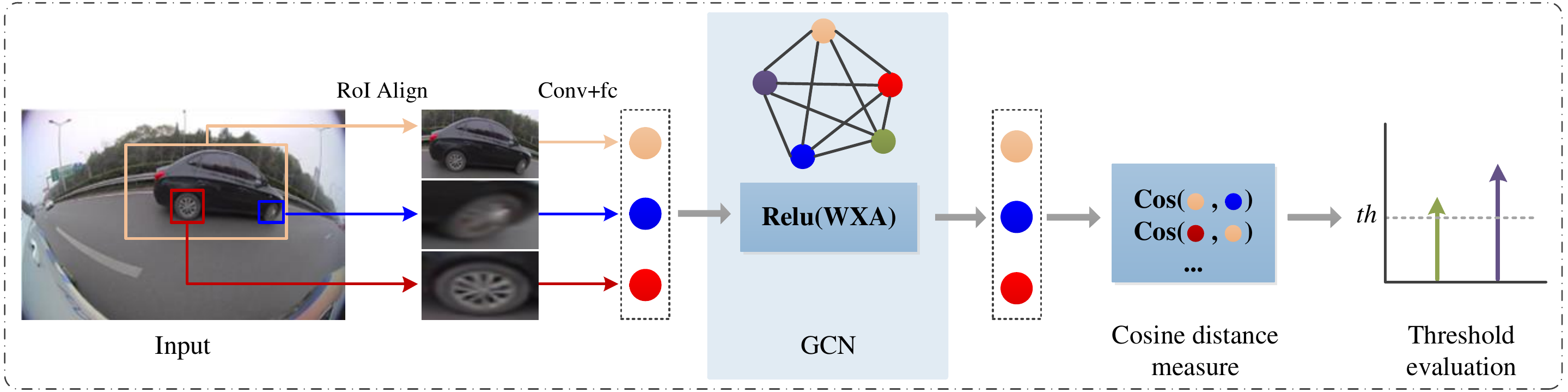}
    \caption{The overall framework of the proposed DeepWORD. The input is the detected bounding box, and after ROI Align the images are the same size. Whereafter, it generates corresponding feature vectors with MLP operation, learning the owner-member relationship with GCN to update the feature vectors. Further, we calculate the cosine distance between the feature vectors from the wheels and vehicles, and retain the wheel-vehicle pairs greater than the threshold as the final results.}
    \label{fig:Figure03}
\end{figure*}

In this paragraph, we present all the details included in the method, such as the overall structure of the model, the GCN module, the principle of GAT.

\subsection{Overall framework of the proposed method}

The overall framework of the proposed DeepWORD is shown in Figure \ref{fig:Figure03}, which consists of two parts, the detection network and the relationship prediction network. The detection network is CenterNet \cite{article24}, whose main task is to detect the vehicle vehicle and tires from the images, and input the detection results into the relationship prediction network. The final model will output the owner-membership relationship between the wheels and the vehicle obtained by prediction.

In the model, we use two Gaussian mixture distributions from prior statistics to predict the initial values of the edges of the connected nodes in the network, which represent the degree of association between the nodes. To improve the accuracy of these association, i.e., to reduce the bias of the prior statistics, we introduce GAT in the GCN to dynamically update these edges. The feature vector of each object is updated once after GCN modul, and GAT associates each object with more global semantic information, thus improving the association between the vehicle and wheels belonging to the same vehicle. Finally, we use cosine distance to measure the similarity between vehicle and wheels, i.e., the probability of belonging to the same vehicle, and keep the pairs with a threshold above 0.5.

\begin{figure}[!t]
    \centering
    \includegraphics[width=5cm]{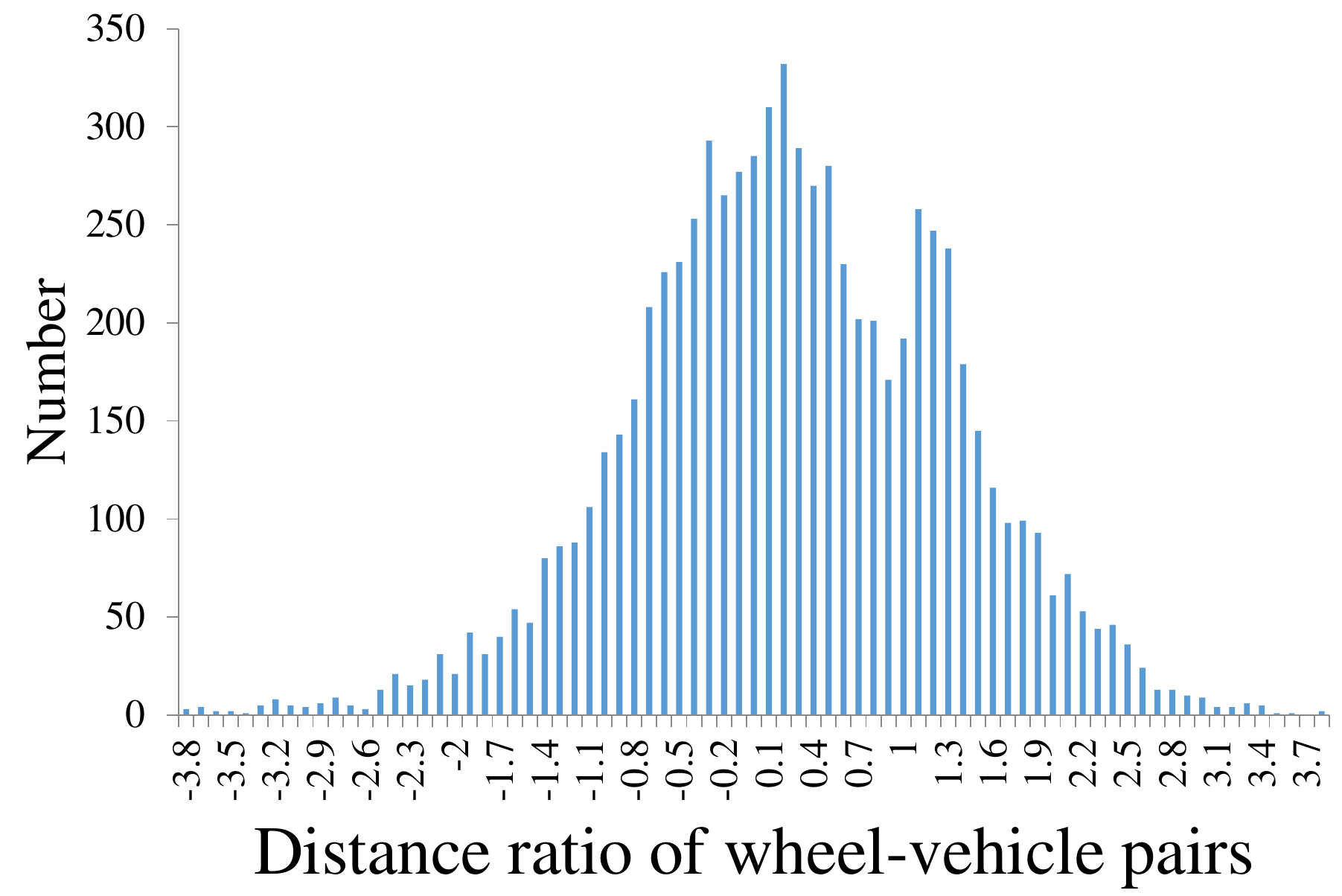}
    \caption{The distance ratio distribution of wheel-vehicle pairs.}
    \label{fig:Figure06}
\end{figure}

\subsection{GCN-based relationship prediction network}

In this section, we explain each part of the GCN-based relationship prediction network.

\paragraph{Prior statistical relationship} The use of a priori statistical relationships between wheels and vehicles has a significant positive effect on the update of nodes in GCN. The relationship between wheels and vehicles after Gaussian distribution is shown in Figure \ref{fig:Figure06}. Obviously, there is a spatially significant interrelationship between the bounding boxes of wheels and vehicle belonging to the same vehicle. For example, for most cases, the wheels belonging to the vehicle are at a shorter distance from the vehicle than the wheels of other vehicles, and the wheels are usually in the lower half of the vehicle's bounding box.

As stated above, we conducted a statistical analysis of the distance ratio of wheel-wheel and wheel-vehicle pairs. To eliminate the interference of different sizes of objects, we normalise them as follows:

\begin{equation}
    \centering 
    D^2  = (\frac{A_\text{j}}{W}-\frac{B_\text{j}}{W})^2+(\frac{A_\text{i}}{H}-\frac{B_\text{i}}{H})^2
    \label{Eq(1)}
\end{equation} 

\begin{figure*}[thpb]
    \centering
    \includegraphics[width=0.95\textwidth]{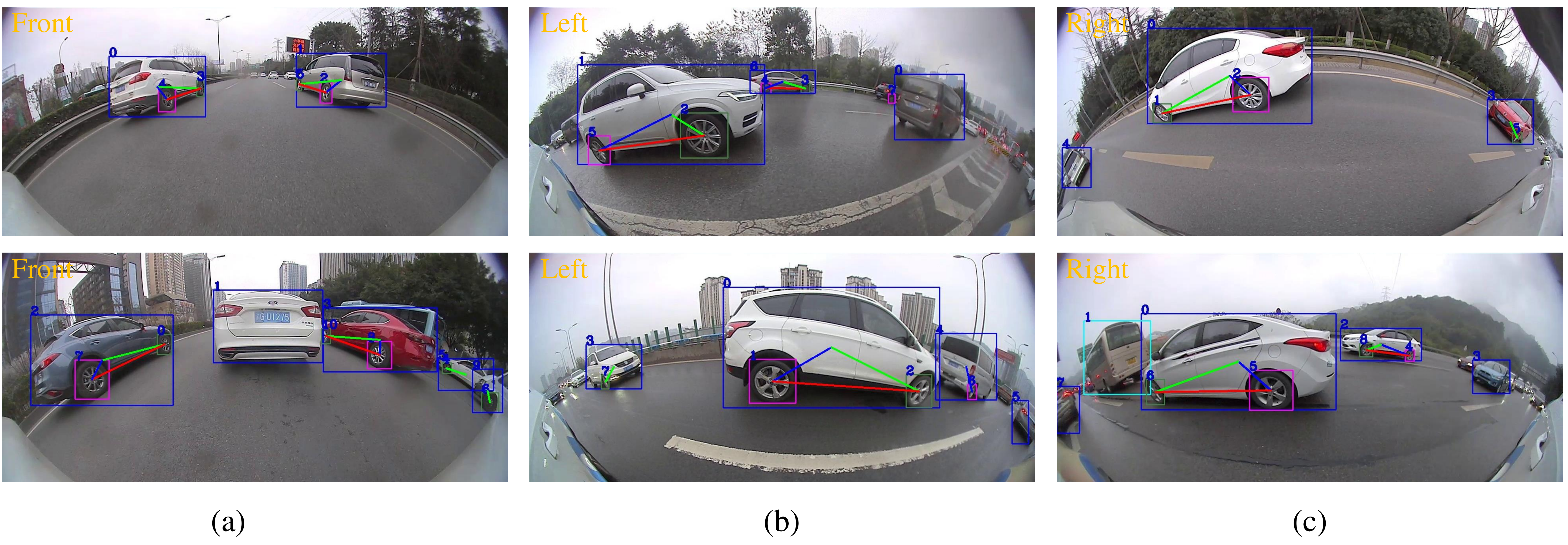}
    \caption{The visualization of the owner-member relationship between wheels and vehicles. (a) Front camera; (b) Left camera; (c) Right camera. Where the upper left corner  is the number of each object, the red line between two wheels means that they are a couple. The green and blue lines respectively connect the rear and front wheels and the vehicles that they belong to.}
    \label{fig:Figure04}
\end{figure*}
\noindent

where $A$ denotes the vehicle and $B$ denotes the wheel in the wheel-vehicle pair, whereas in the wheel-wheel pair, $A$ is the rear wheel, and $B$ is the front wheel. $A_\text{j}$ and $B_\text{j}$ represent the horizontal positions of $A$ and $B$, and $A_\text{i}$ and $B_\text{i}$ refer to the vertical positions of $A$ and $B$, respectively. $W$ and $H$ are the width and height of the input image, respectively.

Distortion of the image and the change of object scale from far to near can seriously affect the prediction results, but by using \ref{Eq(2)} to obtain the distance ratio between the vehicle and the wheels. We found that the priori statistics are more consistent with the distribution of the Gaussian mixture model after logarithmic transformation. Further, we apply this data processing method to the matching of front and rear wheel pairs. Thus, the data after two Gaussian mixture models, i.e., the distance ratio, is used as the prior statistic of the adjacency matrix in the GCN to greatly improve the accuracy of the relationship matching

\begin{equation}
Ratio = \frac{2D}{W_\text{B}+H_\text{B}}
\label{Eq(2)}
\end{equation}
\noindent
where $W_\text{B}$ and $H_\text{B}$ represent the width and height of $B$, respectively.

\paragraph{GCN structure} In the proposed GCN structure, we utilise the feature vectors obtained by convolutional layers plus fully connected layers (\textbf{Conv+fc}) as nodes; therefore, our model is more adept at expressing local spatial information. As mentioned earlier a priori statistical relations are used to initialize the adjacency matrix. Then the vehicle and wheel pairings updated by the GCN structure are computed using the cosine distance. The final pairings with scores greater than a threshold are retained, and in our experiments we set the threshold to 0.5

\paragraph{GAT module} Given that the prior statistics are greatly related to the number of samples and the scene richness, we introduce the GAT into GCN to amend the deviation caused by the limited data. Specifically, we can weight GAT linearly for each edge and secondly refine the weights of the edges in GCN, which can alleviate the impact of noise in the available dataset and enhance the representation ability of the network. 

We set the node vector as $\textbf{h}=\left \{ {\overrightarrow{h}_{\text{1}}},{\overrightarrow{h}_{\text{2}}},...,{\overrightarrow{h}_{\text{N}}}\right \}$, where ${\overrightarrow{h}_{\text{i}}}\in R^{F}$, $N$ represents the number of nodes, and $F$ is the number of features in each node. In the GCN structure, we input the features of each node $\overrightarrow{h}$ and their adjacent nodes $\overrightarrow{h}_{\text{i}}$ into the GAT module, and concatenate them through two fully connected layers. Then, we extract the weight matrix $\textbf{W}= \left \{ w_{\text{1}},w_{\text{2}},...,w_{N\times N} \right \}$ by fully connected layers (FC) and nonlinear activation layers, where $w_{\text{i}}\in R^{F\times F}$.

\setlength\abovedisplayskip{5pt}
\setlength\belowdisplayskip{5pt}
\begin{equation}
Net(X_\text{1},X_\text{2}) \rightarrow FC(Relu(FC(Concat(X_\text{1},X_\text{2}))))
\label{Eq(3)}
\end{equation}
\noindent

where $Net$ denotes the procedure for dealing with the GCN, $(X_{\text{1}},X_{\text{2}})$ represents the features of each node with their adjacent nodes, and $FC$ is a fully connected layer.

Following, we utilize the softmax function to normalize the output and obtain the attention coefficient $Scale_i$:

Next, we utilise the softmax function to normalise the output and obtain the attention coefficient $Scale_i$:

\begin{equation}
Scale_{\text{i}}=\frac{exp(Net(\overrightarrow{h},\overrightarrow{h}_{\text{i}}))}{\sum _{\text{j}}exp(Net(\overrightarrow{h},\overrightarrow{h}_{\text{i}}))}
\label{Eq(4)}
\end{equation}
Finally, we update the original edge weight $w_{\text{i}}$ between the node and the adjacent node by multiplying it with $Scale_{\text{i}}$ to obtain a new weight $w_{\text{i}}^{'}$ as follows:

\setlength\abovedisplayskip{5pt}
\setlength\belowdisplayskip{5pt}
\begin{equation}
w_{\text{i}}^{'} = w_{\text{i}}\times Scale_{\text{i}}
\label{Eq(5)}
\end{equation}

In this manner, we can dynamically correct the prior estimation deviation of the edge during the training process, and improve the prediction accuracy of the owner-member relationship between wheels and vehicles.

\paragraph{Label preparation} We adjust the obtained feature maps of the vehicle and wheels to $H\times W$ and normalize them by [0, 1] to obtain the labels of the relationship prediction network. The four coordinates of the object become a matrix of size $H\times W$ after normalization. We concatenate the coordinate matrix with the wheel and vehicle matrices to obtain $H\times W\times \text{7}$ and input it into the relationship network. The Gaussian mixture distribution is created to model the available data so that we can calculate the probabilities of wheel-vehicle pairs and wheel-wheel pairs to generate the initial adjacency matrix. In the adjacency matrix we adjust the values of unwanted objects to zero (e.g. small objects)

\section{EXPERIMENTS}
\subsection{Dataset overview}
We constructed a dataset called WORD (Wheel and Vehicle Owner-Member Relationship Dataset) as a benchmark. WORD mainly covers two typical scenarios of autonomous driving: parking lots and highways. It contains about 9,000 images collected by a surround-view camera system consisting of surround-view fisheye cameras. Where the frame ids of the images represent the order in which they were taken, and images with the same frame id are taken by different fisheye cameras at the same time.

\subsection{Improvements on embedded platforms}

To deploy our model to the embedded platform of an autonomous driving system, we analysed the Qualcomm SNPE acceleration library and found that some operations in the proposed model were not supported. These operations mainly included RoI Align, GCN matrix multiplication operations, input of multiple heads and sizes, etc. Therefore, we made some improvements to the proposed model as follows:

(1) We removed the RoI Align and resized the feature map extracted from the original image to a fixed size ($\text{56} \times \text{56}$).

(2) We used a fully connected layer instead of a matrix multiplication operation in the GCN.

(3) To solve the problems arising from the input of multiple heads and sizes, we removed the FC encoding module of the coordinate value and concatenated it with the cropped vehicles and wheels from the original image to generate the input features of size $H \times W \times \text{7}$.

\subsection{Backbone selection}

We experimented with a series of backbones, as shown in table \ ref {table01}, so that we could better deploy the model into an embedded platform for autonomous driving. Since speed and accuracy are always conflicting, after a trade-off we use the Conv + fc structure as the final backbone choice to extract the feature maps.

\begin{table}
\setlength{\tabcolsep}{6pt}

\caption{Performance comparison of backbone selection. $AP_{v}$ represents the accuracy of image visualization. 10-speed and 1-speed indicate the speed at which the model processes the image when the batch size is 10 and 1, respectively.}
\label{table01}       

\begin{tabular}{lllll}
\hline\noalign{\smallskip}
Backbone & \begin{tabular}[c]{@{}c@{}}Model size\\ (M)\end{tabular} & \begin{tabular}[c]{@{}c@{}}$AP_{v}$\\ (\%)\end{tabular} & \begin{tabular}[c]{@{}c@{}}10-Speed\\ (ms/10imgs)\end{tabular} & \begin{tabular}[c]{@{}c@{}}1-Speed\\ (ms/img)\end{tabular}  \\
\noalign{\smallskip}\hline\noalign{\smallskip}

Conv+fc & \quad68 & 62.83 & \quad28 & \quad5 \\ 
ResNet18 & \quad66 & 92.47 & \quad110 & \quad13 \\ 

\noalign{\smallskip}\hline
\end{tabular}

\end{table}

After determining the backbone we conducted more experiments to ensure the best balance between speed and accuracy, as shown in table \ref {table02}. From the experimental results, we learned that we should instead reduce the number of fully connected layers and increase the depth of the convolutional layers. Finally, after the reduction, the computational size of the model is 28M, but after image visualization the accuracy still reaches 95.7\% of the best performance, but the running speed is twice as fast.

\subsection{Training of the relationship network} 

During the training stage, the Conv+fc structure extracts the features of the input feature matrix. First, the input matrix goes through Conv+fc, and the GCN structure updates these feature vectors. In addition, we normalise the obtained features and calculate the cosine distance of each wheel and vehicle as the final relationship prediction results. Moreover, we use the L2 loss and manually adjust the weights of the positive and negative samples to optimise the model. At the prediction stage, we multiply the predicted matrix with the mask, and the position where the value is greater than 0.5 indicates that the corresponding combination has an owner-member relationship. The mask is primarily used to filter unnecessary objects. The visualisation of the owner-member relationship between wheels and vehicles in highway scenes is shown in Figure \ref{fig:Figure04}.

\begin{table}
\setlength{\tabcolsep}{2.5pt}
\caption{Performance comparison of different parameters setting based on Conv+fc. The measures taken in turn for backbone are: 1) Neg0.1: reduces the weight of negative samples to 0.1; 2) -Neg: decreases the number of negative samples; 3) -Sma: uses a mask to filter too small objects; 4) 56: intercepts the input image size to 56; 5) 56\_ex\_4: deepens 4 convolutions and reduce the amount of fully connected layers.}
\label{table02}       
\begin{tabular}{lllll}
\hline\noalign{\smallskip}
Backbone & \begin{tabular}[c]{@{}c@{}}Model size\\ (M)\end{tabular} & \begin{tabular}[c]{@{}c@{}}$AP_{v}$\\ (\%)\end{tabular} & \begin{tabular}[c]{@{}c@{}}10-Speed\\ (ms/10imgs)\end{tabular} & \begin{tabular}[c]{@{}c@{}}1-Speed\\ (ms/img)\end{tabular}  \\
\noalign{\smallskip}\hline\noalign{\smallskip}
Conv+fc & \quad68 & 62.83 & 28 & 5 \\
Conv+fc Neg0.1 & \quad68 & 69.34 & 28 & 5\\
Conv+fc-Neg & \quad68 & 73.42 & 28 & 5 \\
Conv+fc-sma & \quad68 & 89.21 & 28 & 5 \\
Conv+fc 56 & \quad53 & 90.33 & 18 & 4 \\
Conv+fc 56\_ex\_4 & \quad\textbf{28} & \textbf{95.70} &\textbf{13.1} &\textbf{3.2} \\
\noalign{\smallskip}\hline
\end{tabular}
\end{table}

\subsection{Performance evaluation of DeepWORD}

\begin{figure}[!t]
    \centering
    \includegraphics[width=0.45\textwidth]{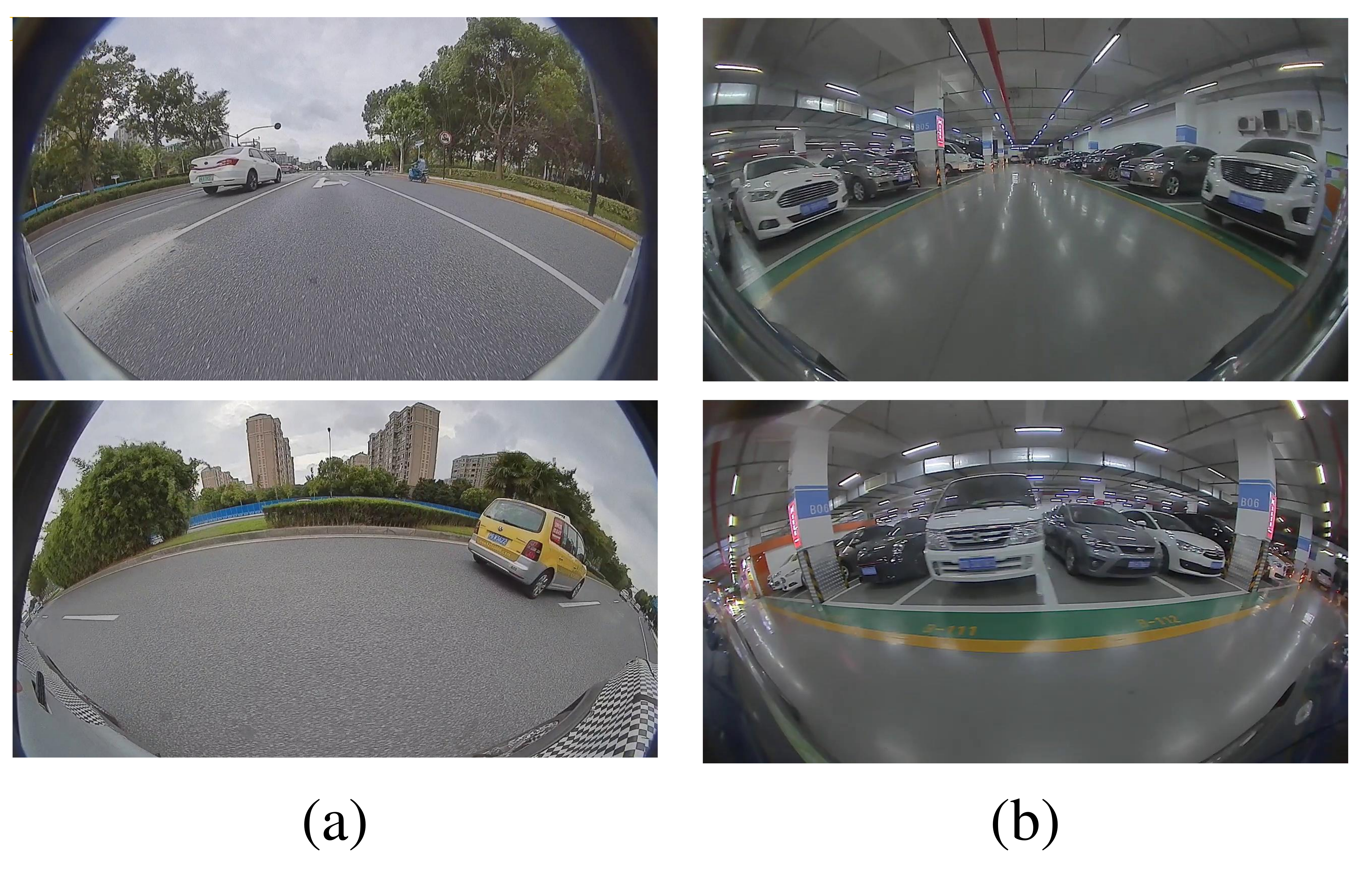}
    \caption{(a) and (b) are easy and hard samples of WORD.}
    \label{fig:Figure05}
\end{figure}

To demonstrate the effectiveness of our structure, we compared the proposed DeepWORD with the previous logic model method. We selected 1000 images to form an easy scene dataset in which each image contained no more than three vehicles, 1000 images to form a difficult scene dataset where each image had more than three vehicles, and a mixed dataset consisting of 500 easy images and 500 difficult images. All samples in these three datasets were randomly selected from the WORD. Some samples are shown in Figure \ref{fig:Figure05}. 

\begin{table}
\setlength{\tabcolsep}{16pt}
\caption{Performance comparison of DeepWORD and logic model method.}
\label{table03}       
\begin{tabular}{llll}
\hline\noalign{\smallskip}
Methods & Easy & Hard & Mixed  \\
\noalign{\smallskip}\hline\noalign{\smallskip}
Logic model & 92.91 & 71.83 & 79.90 \\
DeepWORD &  \textbf{99.17} &  \textbf{94.35} &  \textbf{95.14} \\
\noalign{\smallskip}\hline
\end{tabular}
\end{table}

\begin{figure}[thpb]
   
   \centering
   \subfigure[]{\includegraphics[width=4cm]{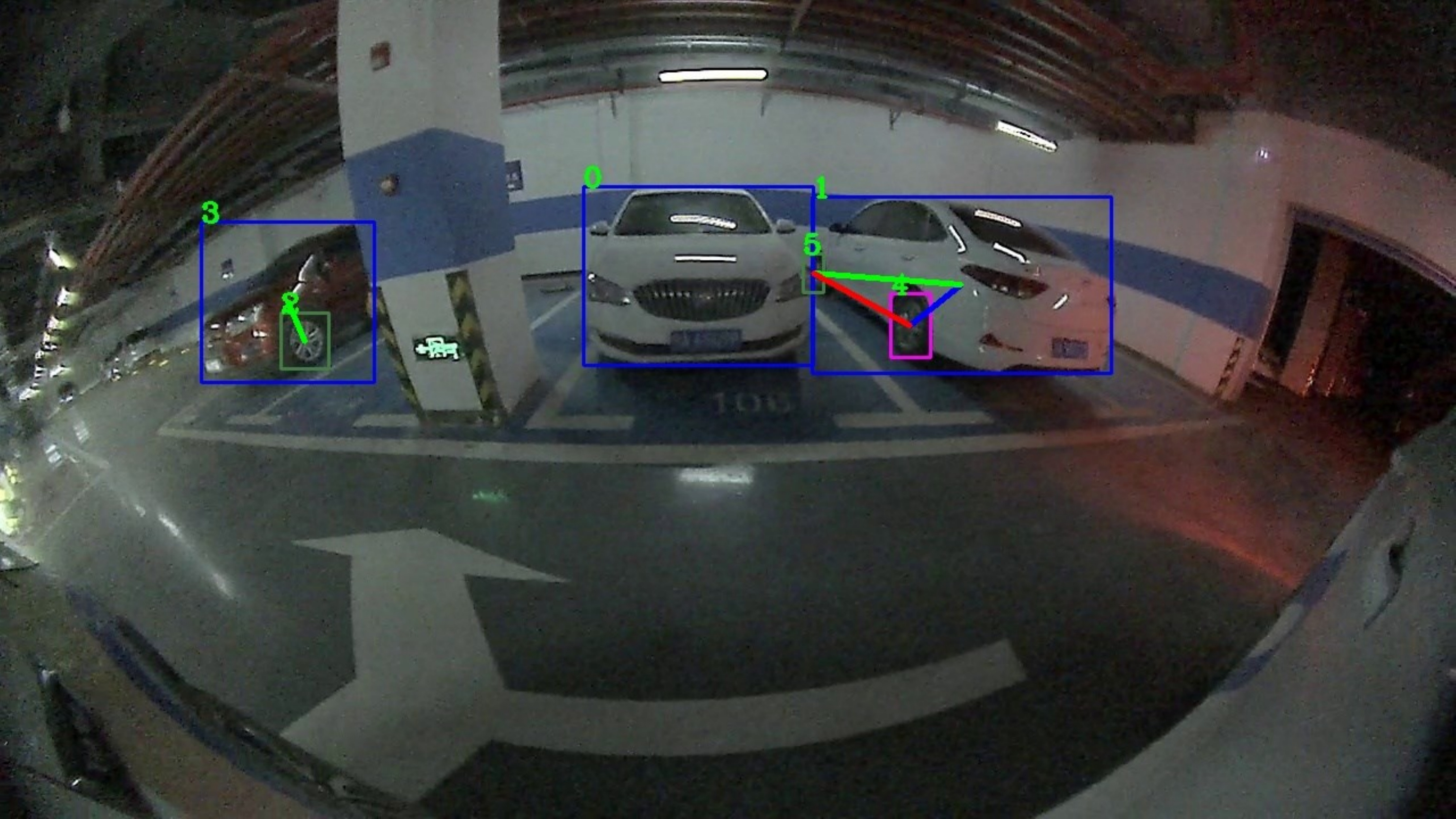}} \hspace{1mm}
   \subfigure[]{\includegraphics[width=4cm]{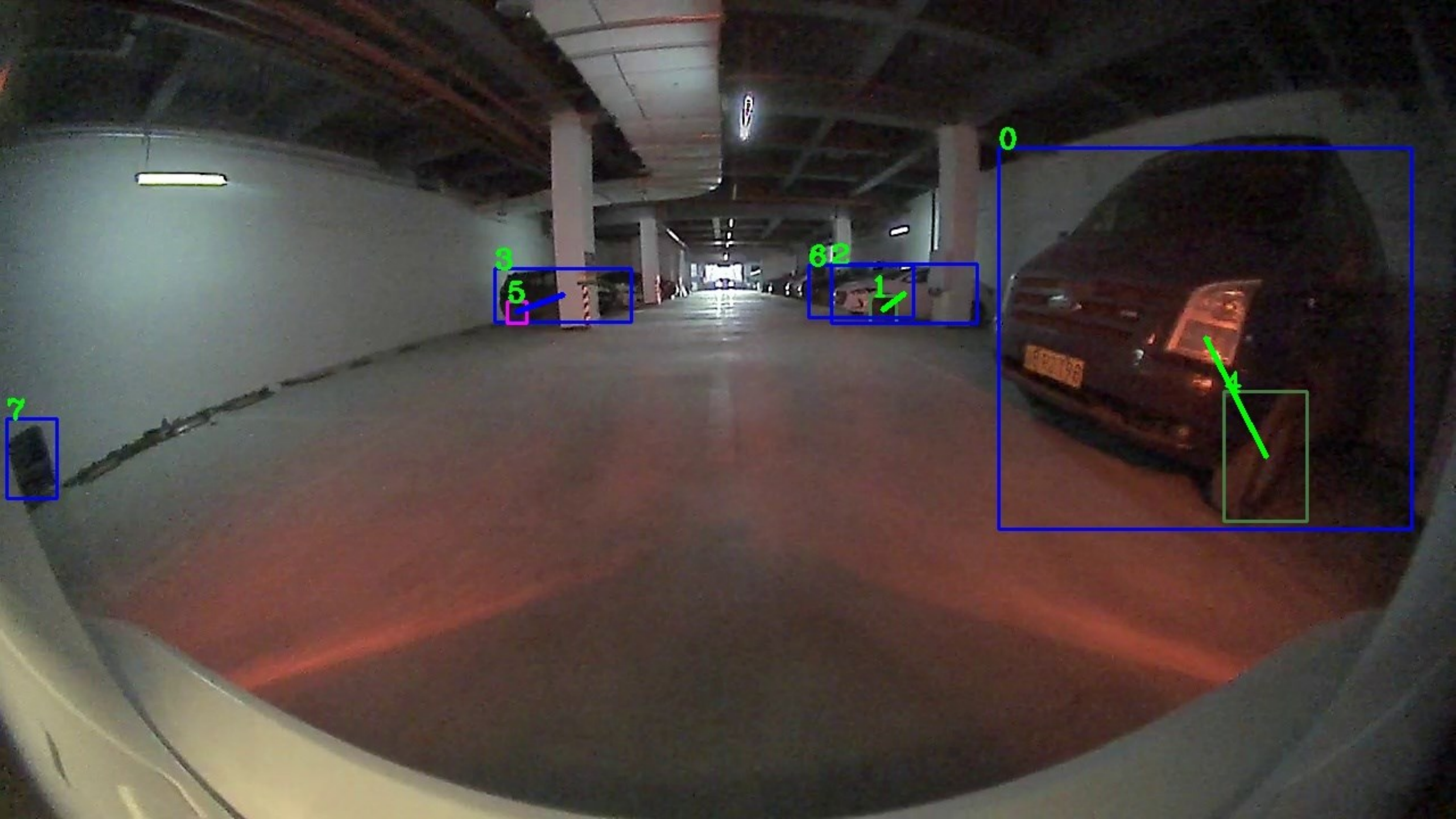}}
   \\ %
   \centering
   \subfigure[]{\includegraphics[width=4cm]{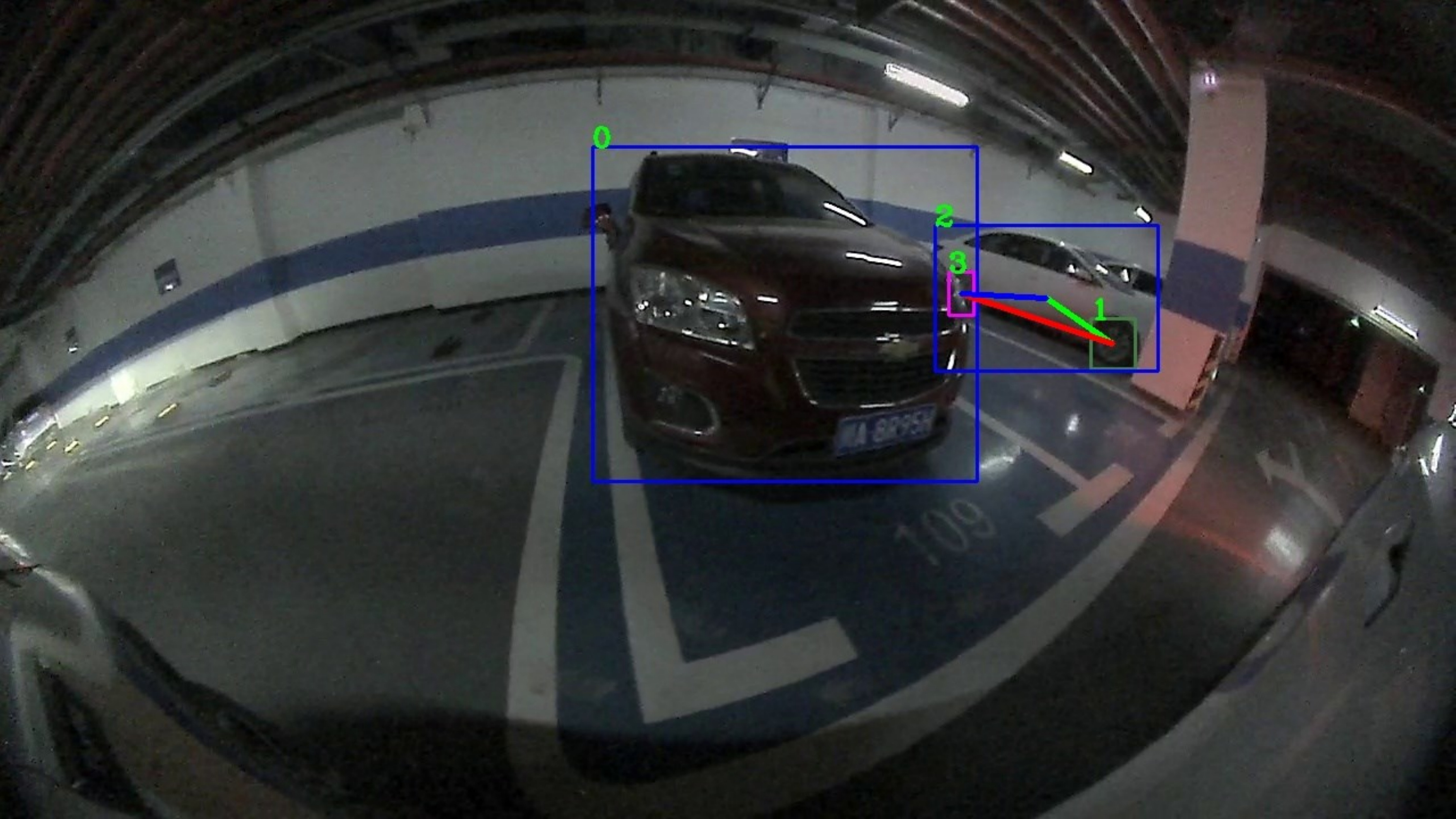}}\hspace{2mm}
   \subfigure[]{\includegraphics[width=4cm]{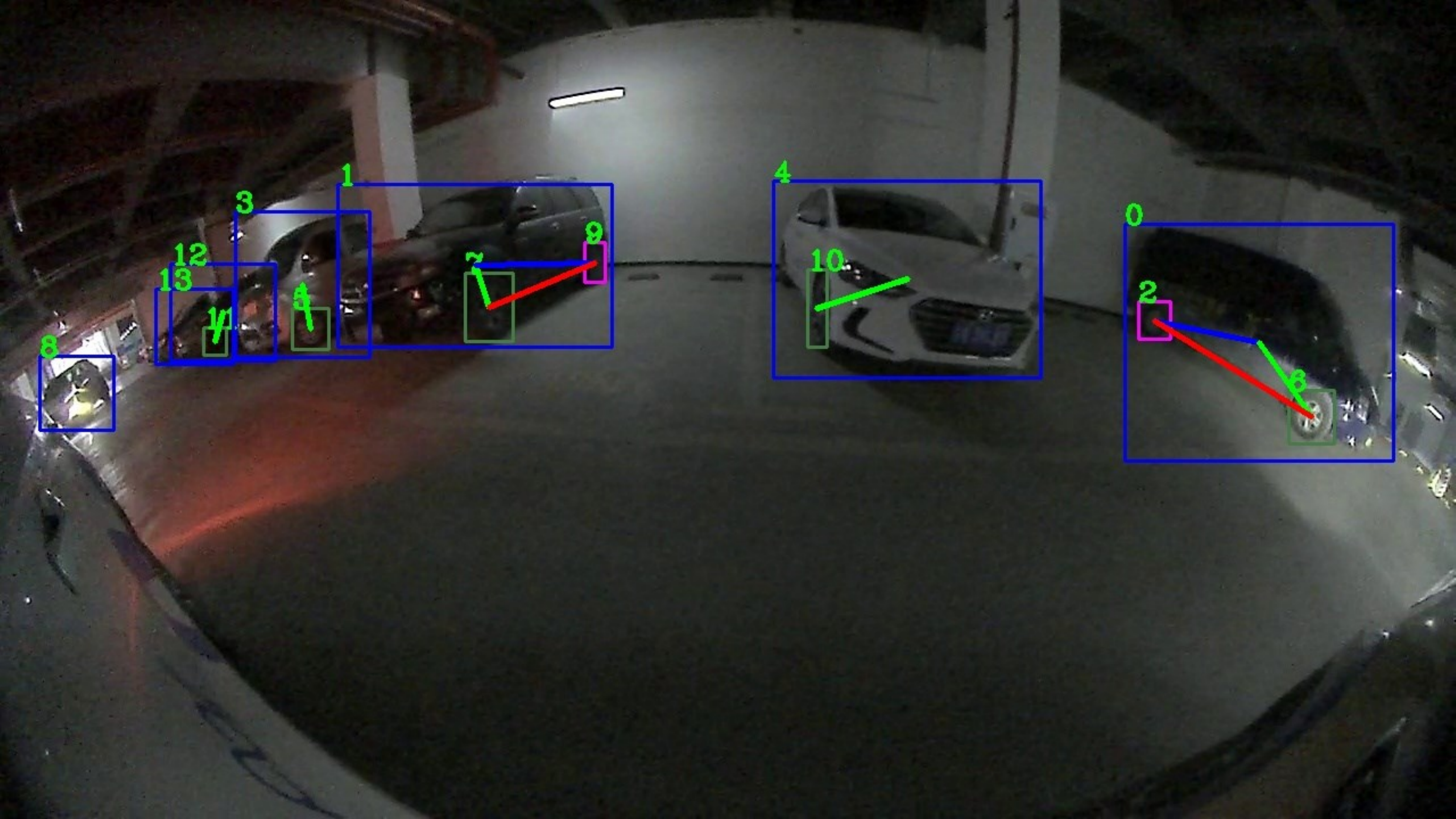}}

   \caption{(a), (b), (c) and (d) are difficult scenes taken by camera while the vehicle is moving in the underground parking lot. The bounding-boxes of vehicles  1 and 0 intersect at wheel 5 as shown in (a),and the bounding-boxes of vehicles  6 and 2 intersect at wheel 1 as shown in (b). The bounding-boxes of vehicles 0 and 2 completely contain wheel 3 as shown in (c), and the bounding-boxes of vehicles 12 and 13 completely contain wheel 11 as shown in (d).}
   \label{fig:Figure07}
\end{figure}

The performance of DeepWORD in simple and difficult scenarios is shown in Figures \ref{fig:Figure04} and \ref{fig:Figure07}, respectively. We find that DeepWORD performs very well in difficult scenes, such as in a parking lot, where cars are parked parallel and close to each other. The owner-member relationship can be correctly identified even if the bounding boxes of two different cars intersect at the same wheel, or the bounding boxes of different vehicles completely contain the same wheel, which is very difficult to solve using the logic model based on IoU.

Finally, we conducted experiments to compare the classification accuracy of these three datasets. As shown in Table \ref{table03}, our DeepWORD model improves the accuracy by 6.26\% in simple scenarios when compared with the traditional logic model approach. In contrast, DeepWORD performs even better in complex scenarios, with accuracy improvements of 22.52\% and 15.24\%, respectively. After observing the prediction results, the traditional logic model is more likely to predict the wrong owner-member relationship when the vehicles appear to be densely arranged. It is worth noting that both the generalization ability and prediction accuracy of DeepWORD perform excellent relative to the traditional logic model approach.

\section{CONCLUSION AND FUTURE WORK}

In this paper, our main research result is a GCN-based owner-membership relationship prediction network DeepWORD for complex scenarios, always with good applicability and prediction accuracy. Moreover, we apply it to vehicle-mounted surround-view camera systems and conclude that it is efficient and effective after extensive experiments. In addition, to promote related research, we have established and released the large-scale relational dataset WORD as the first dataset for autonomous driving in the field of relation prediction. In the future, we will continue to extend the WORD dataset for more complex real-world scenarios and further optimize our owner-member relationship prediction solution DeepWORD.

\bibliographystyle{cas-model2-names}


\begin{thebibliography}{23}
\expandafter\ifx\csname natexlab\endcsname\relax\def\natexlab#1{#1}\fi
\providecommand{\url}[1]{\texttt{#1}}
\providecommand{\href}[2]{#2}
\providecommand{\path}[1]{#1}
\providecommand{\DOIprefix}{doi:}
\providecommand{\ArXivprefix}{arXiv:}
\providecommand{\URLprefix}{URL: }
\providecommand{\Pubmedprefix}{pmid:}
\providecommand{\doi}[1]{\href{http://dx.doi.org/#1}{\path{#1}}}
\providecommand{\Pubmed}[1]{\href{pmid:#1}{\path{#1}}}
\providecommand{\bibinfo}[2]{#2}
\ifx\xfnm\relax \def\xfnm[#1]{\unskip,\space#1}\fi
\bibitem[{{Antonelli} et~al.(1999){Antonelli}, {Chiaverini}, {Finotello} and
  {Morgavi}}]{article19}
\bibinfo{author}{{Antonelli}, G.}, \bibinfo{author}{{Chiaverini}, S.},
  \bibinfo{author}{{Finotello}, R.}, \bibinfo{author}{{Morgavi}, E.},
  \bibinfo{year}{1999}.
\newblock \bibinfo{title}{Real-time path planning and obstacle avoidance for an
  autonomous underwater vehicle}, in: \bibinfo{booktitle}{Proceedings 1999 IEEE
  International Conference on Robotics and Automation (Cat. No.99CH36288C)}.
\bibitem[{{Arrospide} et~al.(2010){Arrospide}, {Salgado}, {Nieto} and
  {Mohedano}}]{article20}
\bibinfo{author}{{Arrospide}, J.}, \bibinfo{author}{{Salgado}, L.},
  \bibinfo{author}{{Nieto}, M.}, \bibinfo{author}{{Mohedano}, R.},
  \bibinfo{year}{2010}.
\newblock \bibinfo{title}{Homography-based ground plane detection using a
  single on-board camera}.
\newblock \bibinfo{journal}{IET Intelligent Transport Systems} .
\bibitem[{Bruna et~al.(2013)Bruna, Zaremba et~al.}]{article7}
\bibinfo{author}{Bruna, J.}, \bibinfo{author}{Zaremba, W.}, et~al.,
  \bibinfo{year}{2013}.
\newblock \bibinfo{title}{Spectral networks and locally connected networks on
  graphs}.
\newblock \bibinfo{journal}{arXiv preprint arXiv:1312.6203} .
\bibitem[{{Chu} et~al.(2018){Chu}, {Gong}, {Shao}, {Chang} and
  {Ni}}]{article21}
\bibinfo{author}{{Chu}, H.}, \bibinfo{author}{{Gong}, K.},
  \bibinfo{author}{{Shao}, Y.}, \bibinfo{author}{{Chang}, Z.},
  \bibinfo{author}{{Ni}, J.}, \bibinfo{year}{2018}.
\newblock \bibinfo{title}{3d perception and reconstruction system based on 2d
  laser scanner}, in: \bibinfo{booktitle}{2018 Chinese Automation Congress
  (CAC)}.
\bibitem[{Dai et~al.(2017)Dai, Zhang and Lin}]{article4}
\bibinfo{author}{Dai, B.}, \bibinfo{author}{Zhang, Y.}, \bibinfo{author}{Lin,
  D.}, \bibinfo{year}{2017}.
\newblock \bibinfo{title}{Detecting visual relationships with deep relational
  networks}, in: \bibinfo{booktitle}{Proceedings of the IEEE conference on
  computer vision and Pattern recognition}, pp. \bibinfo{pages}{3076--3086}.
\bibitem[{Defferrard et~al.(2016)Defferrard, Bresson et~al.}]{article8}
\bibinfo{author}{Defferrard, M.}, \bibinfo{author}{Bresson, X.}, et~al.,
  \bibinfo{year}{2016}.
\newblock \bibinfo{title}{Convolutional neural networks on graphs with fast
  localized spectral filtering}, in: \bibinfo{booktitle}{Advances in neural
  information processing systems}, pp. \bibinfo{pages}{3844--3852}.
\bibitem[{Desai et~al.(2011)Desai, Ramanan and Fowlkes}]{article3}
\bibinfo{author}{Desai, C.}, \bibinfo{author}{Ramanan, D.},
  \bibinfo{author}{Fowlkes, C.C.}, \bibinfo{year}{2011}.
\newblock \bibinfo{title}{Discriminative models for multi-class object layout}.
\newblock \bibinfo{journal}{International journal of computer vision}
  \bibinfo{volume}{95}, \bibinfo{pages}{1--12}.
\bibitem[{Gregor et~al.(2002)Gregor, Lutzeler, Pellkofer et~al.}]{article15}
\bibinfo{author}{Gregor, R.}, \bibinfo{author}{Lutzeler, M.},
  \bibinfo{author}{Pellkofer, M.}, et~al., \bibinfo{year}{2002}.
\newblock \bibinfo{title}{Ems-vision: A perceptual system for autonomous
  vehicles}.
\newblock \bibinfo{journal}{Intelligent Transportation Systems, IEEE
  Transactions on} .
\bibitem[{Grigorescu et~al.(2020)Grigorescu, Trasnea et~al.}]{article13}
\bibinfo{author}{Grigorescu, S.}, \bibinfo{author}{Trasnea, B.}, et~al.,
  \bibinfo{year}{2020}.
\newblock \bibinfo{title}{A survey of deep learning techniques for autonomous
  driving}.
\newblock \bibinfo{journal}{Journal of Field Robotics} .
\bibitem[{Hamilton et~al.(2017)Hamilton, Ying and Leskovec}]{article10}
\bibinfo{author}{Hamilton, W.}, \bibinfo{author}{Ying, Z.},
  \bibinfo{author}{Leskovec, J.}, \bibinfo{year}{2017}.
\newblock \bibinfo{title}{Inductive representation learning on large graphs},
  in: \bibinfo{booktitle}{Advances in neural information processing systems},
  pp. \bibinfo{pages}{1024--1034}.
\bibitem[{Häne et~al.(2017)Häne, Heng,  et~al.}]{article14}
\bibinfo{author}{Häne, C.}, \bibinfo{author}{Heng, L.}, , et~al.,
  \bibinfo{year}{2017}.
\newblock \bibinfo{title}{3d visual perception for self-driving cars using a
  multi-camera system: Calibration, mapping, localization, and obstacle
  detection}.
\newblock \bibinfo{journal}{arXiv preprint arXiv:1708.09839} .
\bibitem[{Javanmardi et~al.(2019)Javanmardi, Gu, Javanmardi and
  Kamijo}]{article17}
\bibinfo{author}{Javanmardi, E.}, \bibinfo{author}{Gu, Y.},
  \bibinfo{author}{Javanmardi, M.}, \bibinfo{author}{Kamijo, S.},
  \bibinfo{year}{2019}.
\newblock \bibinfo{title}{Autonomous vehicle self-localization based on
  abstract map and multi-channel lidar in urban area}.
\newblock \bibinfo{journal}{IATSS Research} .
\bibitem[{Kipf and Welling(2016)}]{article9}
\bibinfo{author}{Kipf, T.N.}, \bibinfo{author}{Welling, M.},
  \bibinfo{year}{2016}.
\newblock \bibinfo{title}{Semi-supervised classification with graph
  convolutional networks}.
\newblock \bibinfo{journal}{arXiv preprint arXiv:1609.02907} .
\bibitem[{Minguez et~al.(2008)Minguez, Lamiraux and Laumond}]{article18}
\bibinfo{author}{Minguez, J.}, \bibinfo{author}{Lamiraux, F.},
  \bibinfo{author}{Laumond, J.P.}, \bibinfo{year}{2008}.
\newblock \bibinfo{title}{Motion planning and obstacle avoidance}.
\newblock \bibinfo{journal}{Springer Handbook of Robotics} .
\bibitem[{Shuman et~al.(2013)Shuman, Narang et~al.}]{article6}
\bibinfo{author}{Shuman, D.I.}, \bibinfo{author}{Narang, S.K.}, et~al.,
  \bibinfo{year}{2013}.
\newblock \bibinfo{title}{The emerging field of signal processing on graphs:
  Extending high-dimensional data analysis to networks and other irregular
  domains}.
\newblock \bibinfo{journal}{IEEE signal processing magazine}
  \bibinfo{volume}{30}, \bibinfo{pages}{83--98}.
\bibitem[{Staudemeyer and Morris(2019)}]{article26}
\bibinfo{author}{Staudemeyer, R.C.}, \bibinfo{author}{Morris, E.R.},
  \bibinfo{year}{2019}.
\newblock \bibinfo{title}{Understanding lstm -- a tutorial into long short-term
  memory recurrent neural networks}.
\newblock \bibinfo{journal}{arXiv preprint arXiv:1909.09586} .
\bibitem[{Sutton and McCallum(2010)}]{article25}
\bibinfo{author}{Sutton, C.}, \bibinfo{author}{McCallum, A.},
  \bibinfo{year}{2010}.
\newblock \bibinfo{title}{An introduction to conditional random fields}.
\newblock \bibinfo{journal}{arXiv preprint arXiv:1011.4088} .
\bibitem[{Veli{\v{c}}kovi{\'c} et~al.(2017)Veli{\v{c}}kovi{\'c}, Cucurull
  et~al.}]{article11}
\bibinfo{author}{Veli{\v{c}}kovi{\'c}, P.}, \bibinfo{author}{Cucurull, G.},
  et~al., \bibinfo{year}{2017}.
\newblock \bibinfo{title}{Graph attention networks}.
\newblock \bibinfo{journal}{arXiv preprint arXiv:1710.10903} .
\bibitem[{Vivet et~al.(2012)Vivet, Checchin and Chapuis}]{article22}
\bibinfo{author}{Vivet, D.}, \bibinfo{author}{Checchin, P.},
  \bibinfo{author}{Chapuis, R.}, \bibinfo{year}{2012}.
\newblock \bibinfo{title}{Radar-only localization and mapping for ground
  vehicle at high speed and for riverside boat}, in:
  \bibinfo{booktitle}{10.1109/ICRA.2012.6224573}.
\bibitem[{Wang et~al.(2017)Wang, Zhang and Wang}]{article16}
\bibinfo{author}{Wang, L.}, \bibinfo{author}{Zhang, Y.}, \bibinfo{author}{Wang,
  J.}, \bibinfo{year}{2017}.
\newblock \bibinfo{title}{Map-based localization method for autonomous vehicles
  using 3d-lidar}.
\newblock \bibinfo{journal}{IFAC-PapersOnLine} .
\bibitem[{Xu et~al.(2019)Xu, Jiang, Liang and Li}]{article12}
\bibinfo{author}{Xu, H.}, \bibinfo{author}{Jiang, C.}, \bibinfo{author}{Liang,
  X.}, \bibinfo{author}{Li, Z.}, \bibinfo{year}{2019}.
\newblock \bibinfo{title}{Spatial-aware graph relation network for large-scale
  object detection}, in: \bibinfo{booktitle}{Proceedings of the IEEE Conference
  on Computer Vision and Pattern Recognition}, pp. \bibinfo{pages}{9298--9307}.
\bibitem[{Zellers et~al.(2018)Zellers, Yatskar et~al.}]{article5}
\bibinfo{author}{Zellers, R.}, \bibinfo{author}{Yatskar, M.}, et~al.,
  \bibinfo{year}{2018}.
\newblock \bibinfo{title}{Neural motifs: Scene graph parsing with global
  context}, in: \bibinfo{booktitle}{Proceedings of the IEEE Conference on
  Computer Vision and Pattern Recognition}, pp. \bibinfo{pages}{5831--5840}.
\bibitem[{Zhou et~al.(2019)Zhou, Wang and Krähenbühl}]{article24}
\bibinfo{author}{Zhou, X.}, \bibinfo{author}{Wang, D.},
  \bibinfo{author}{Krähenbühl, P.}, \bibinfo{year}{2019}.
\newblock \bibinfo{title}{Objects as points}.
\newblock \bibinfo{journal}{arXiv preprint arXiv:1904.07850} .

\end{thebibliography}

\newpage
\appendixpage
\begin{appendices}
      
      \section{\textbf{WORD} Dataset}
\begin{figure}[thpb]

\subfigure[highway scene]{
    \begin{minipage}[b]{0.37\linewidth} 
    \includegraphics[width=4cm, height=2cm]{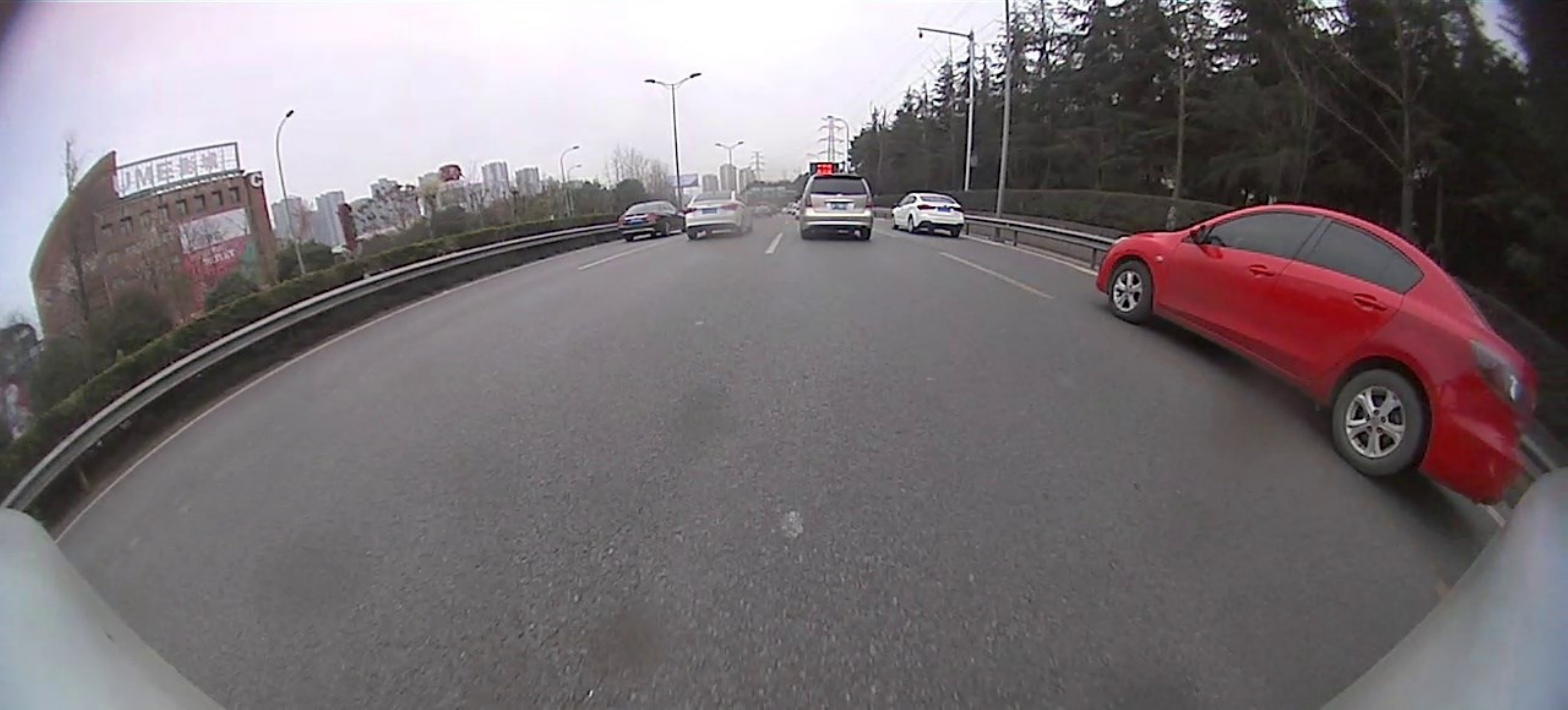}\vspace{1pt} 
    \includegraphics[width=4cm,height=2cm]{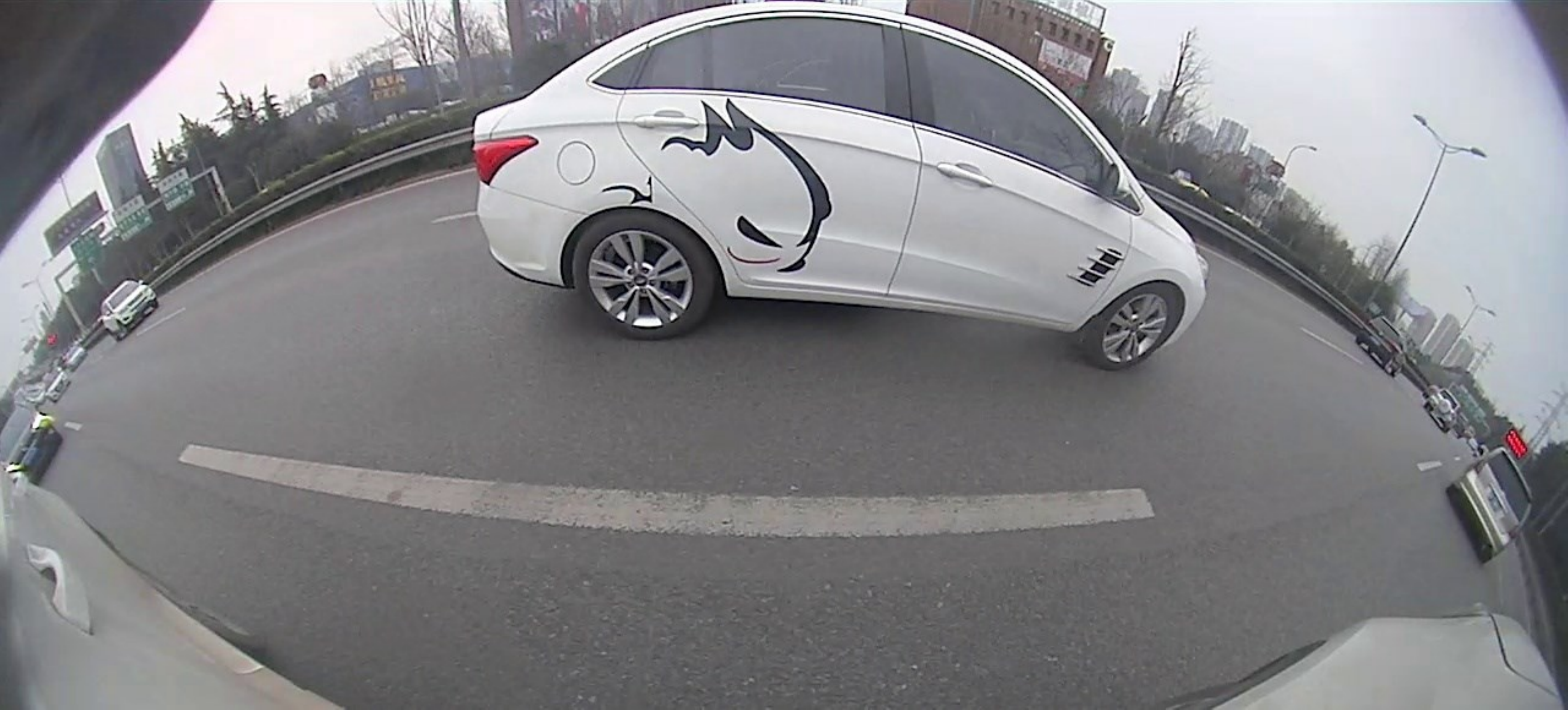}\vspace{1pt}
    \includegraphics[width=4cm,height=2cm]{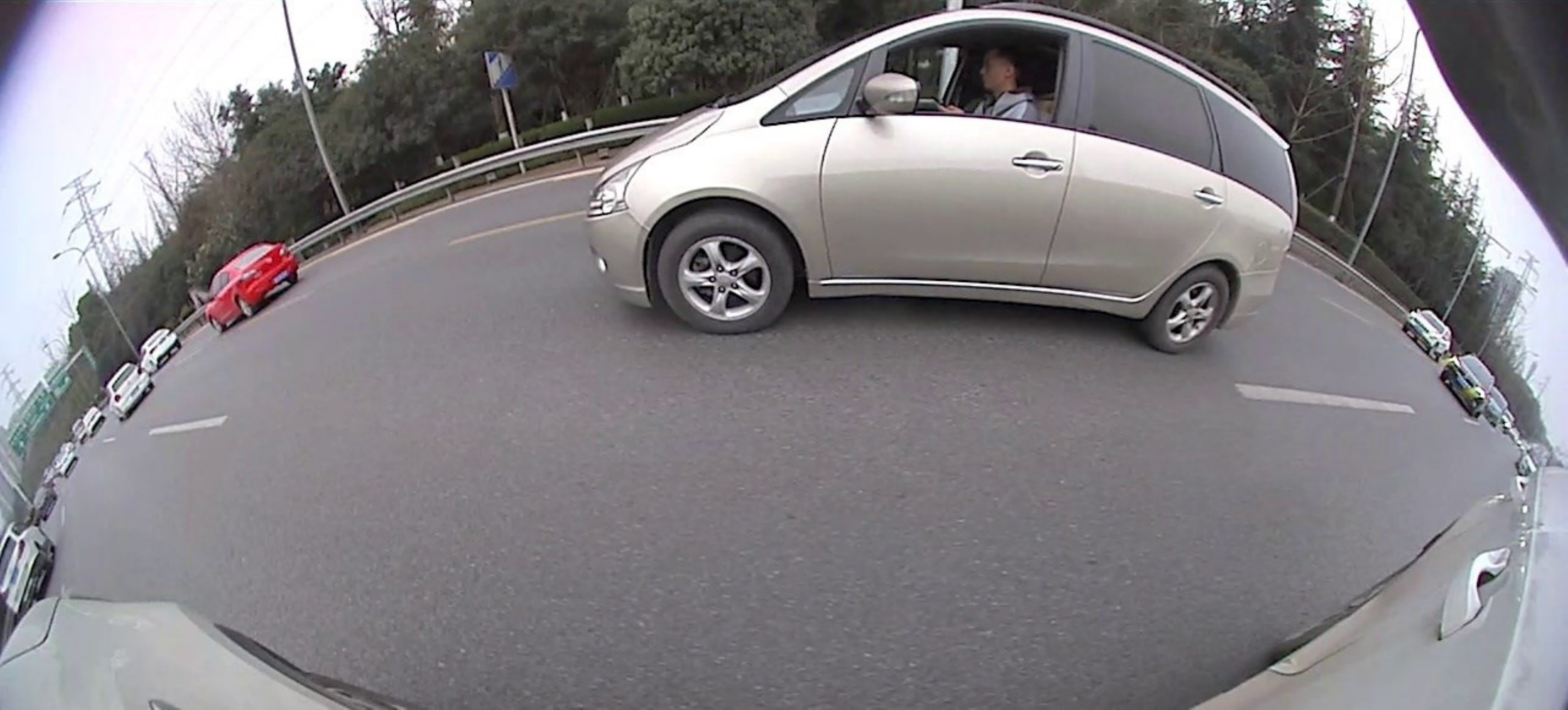}\vspace{1pt}
    \end{minipage}
}
\hspace{7mm}
\subfigure[ parking scene]{
    \begin{minipage}[b]{0.37\linewidth}
    \includegraphics[width=4cm, height=2cm]{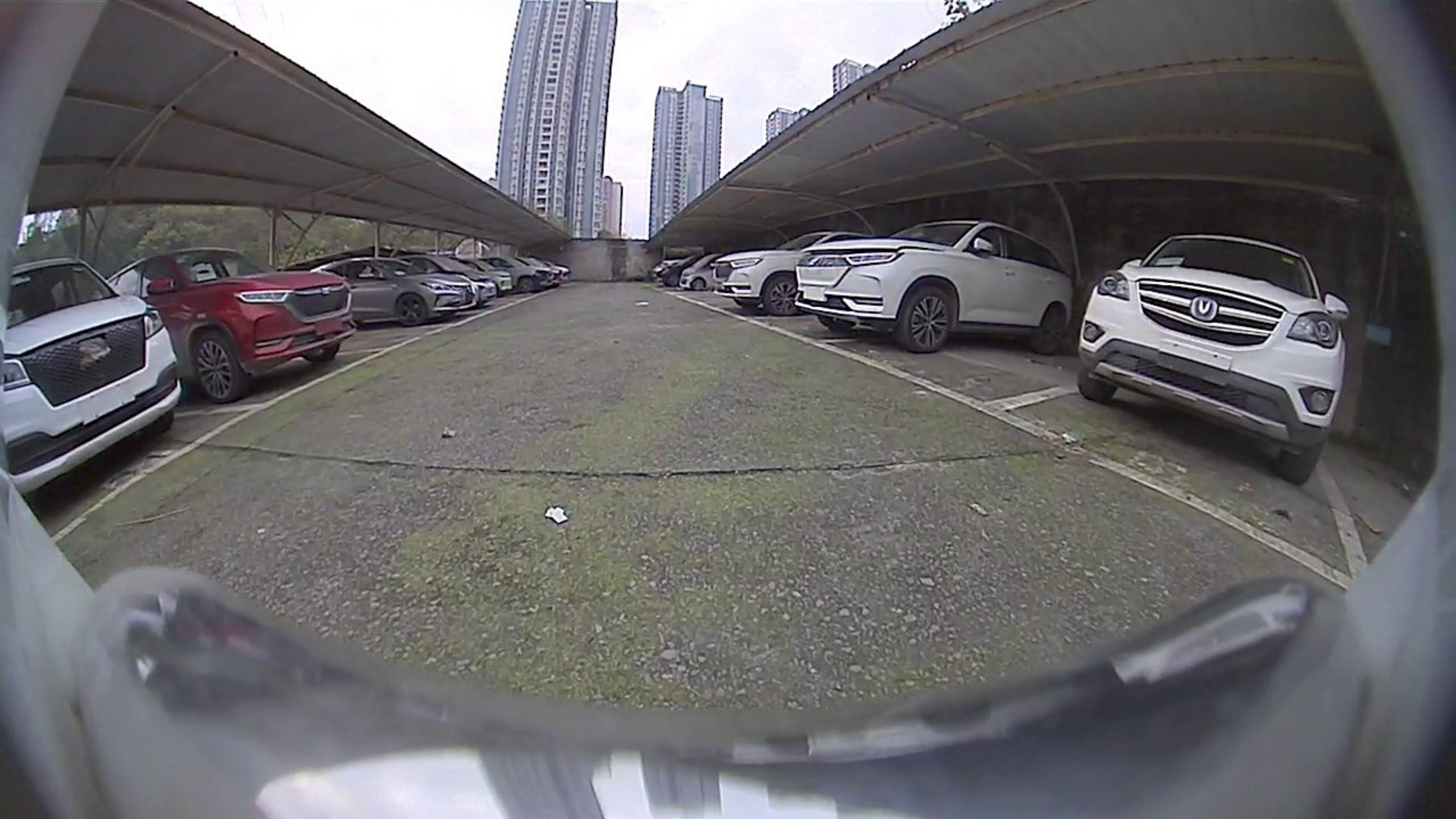}\vspace{1pt} 
    \includegraphics[width=4cm,height=2cm]{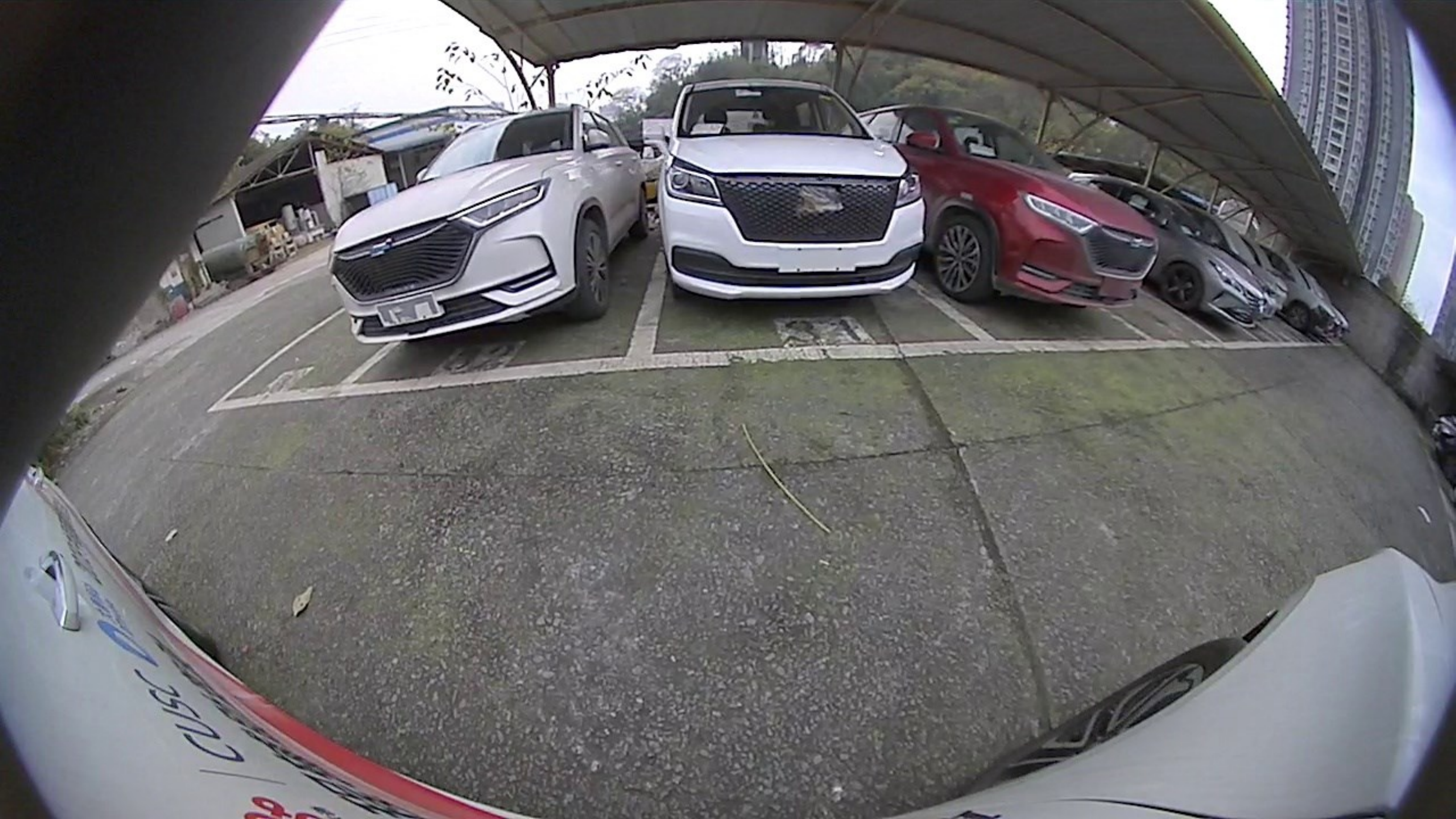}\vspace{1pt}
    \includegraphics[width=4cm,height=2cm]{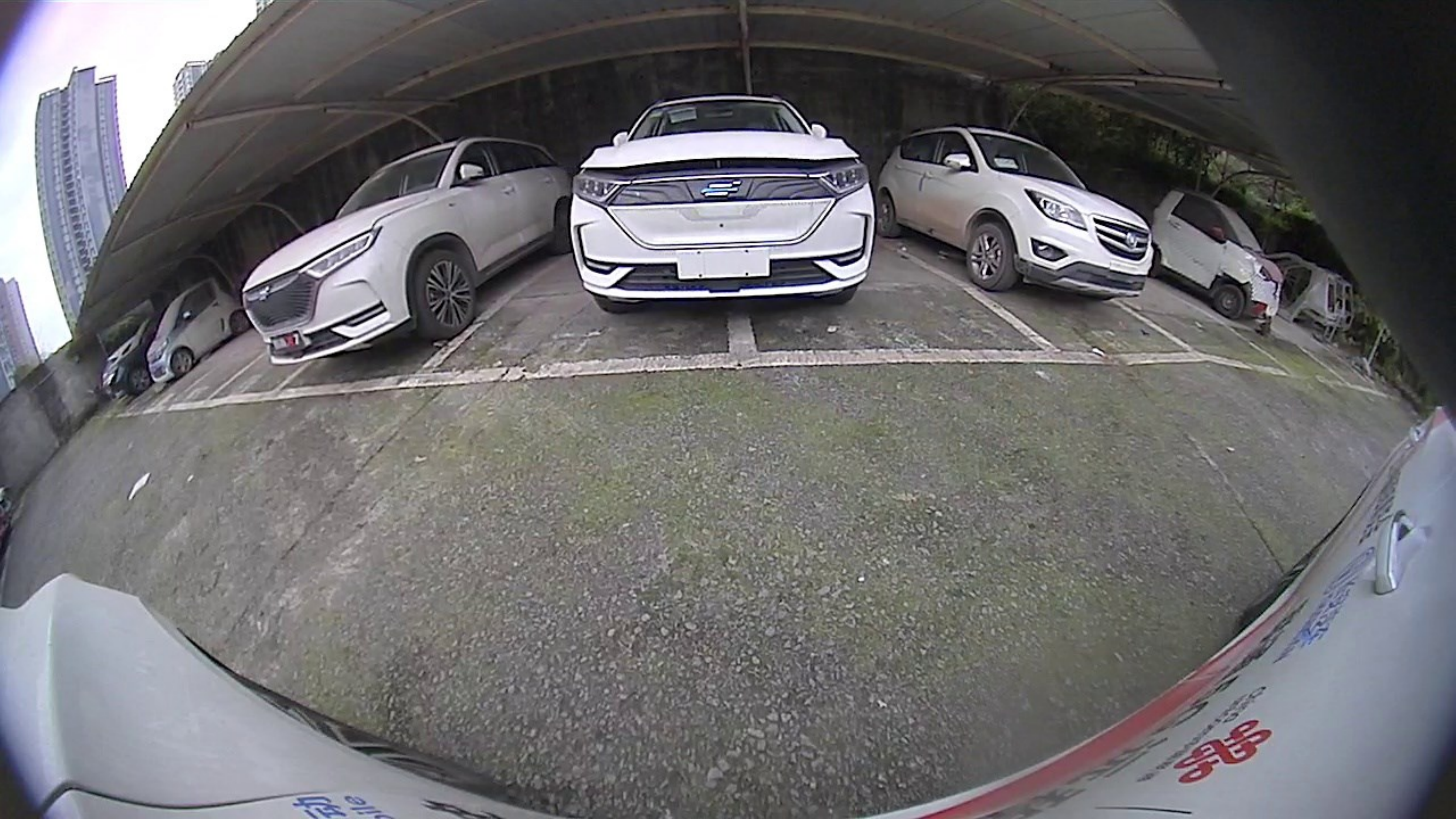}\vspace{1pt}
    \end{minipage}
}
\caption{(a) and (b) show the highway scene and the parking scene. Each scene is captured from top to bottom by the front camera, the left camera and the right camera.}
\label{fig:Figure08}
\end{figure}

      WORD (Wheel and Vehicle Owner-Member Relationship Dataset) contains two categories that are common in autonomous driving: parking scene and highway scene. Examples of these two scenes are shown in Figure \ref{fig:Figure08}.
      
\begin{figure}[!t]
   
   \centering
   \subfigure[front]{\includegraphics[width=4cm]{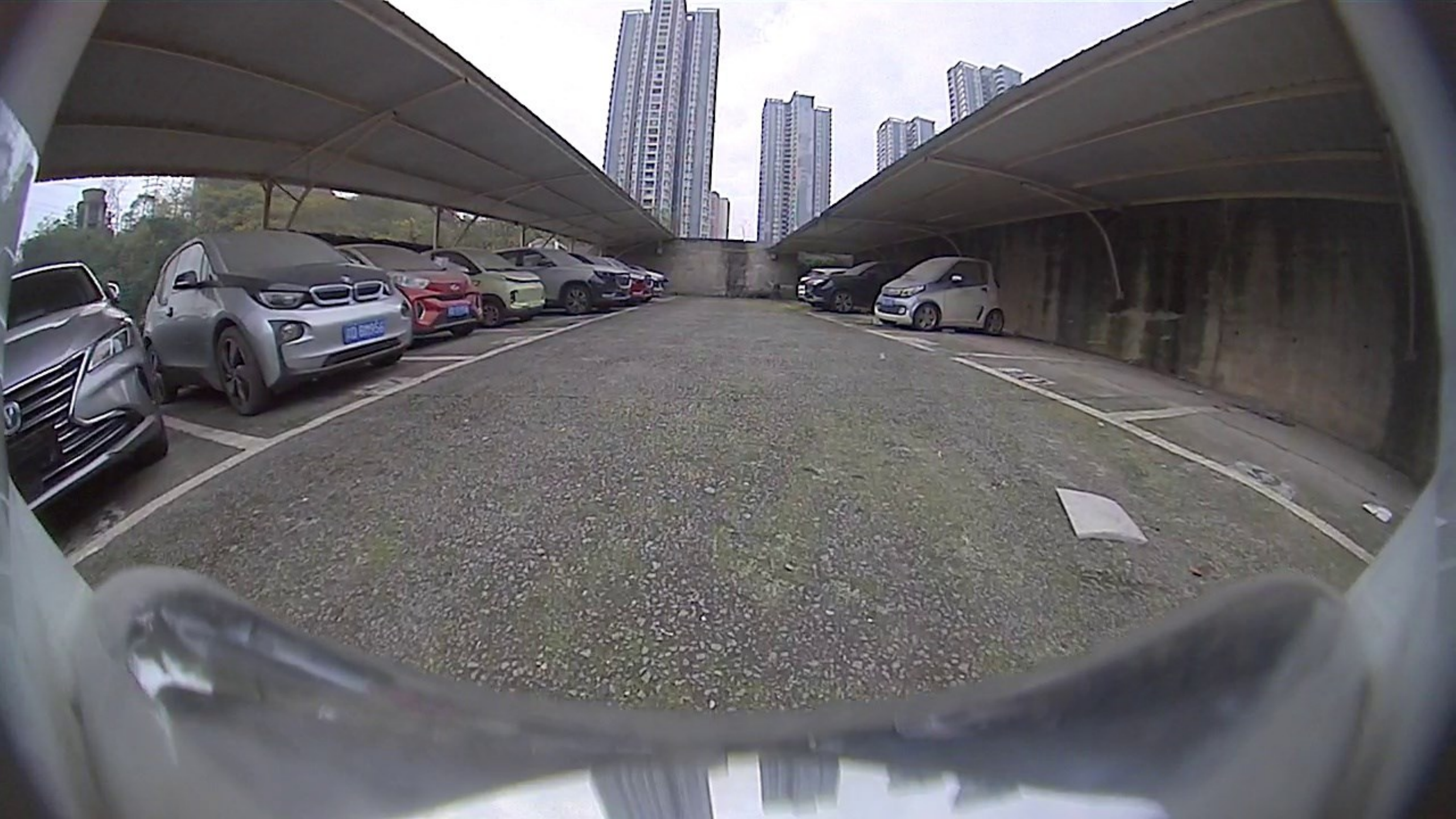}} \hspace{1mm}
   \subfigure[left]{\includegraphics[width=4cm]{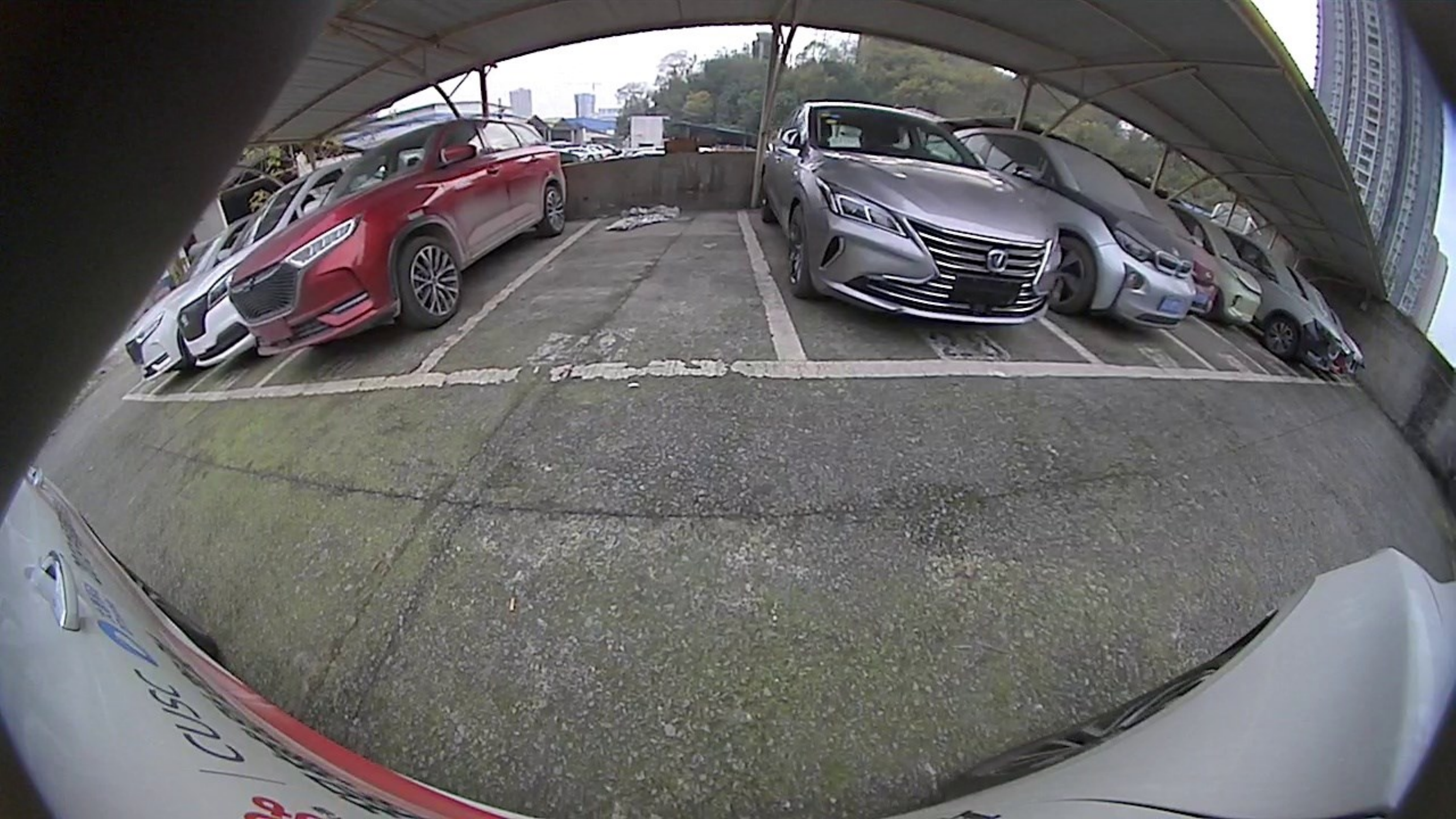}}
   \\ 
   \centering
   \subfigure[right]{\includegraphics[width=4cm]{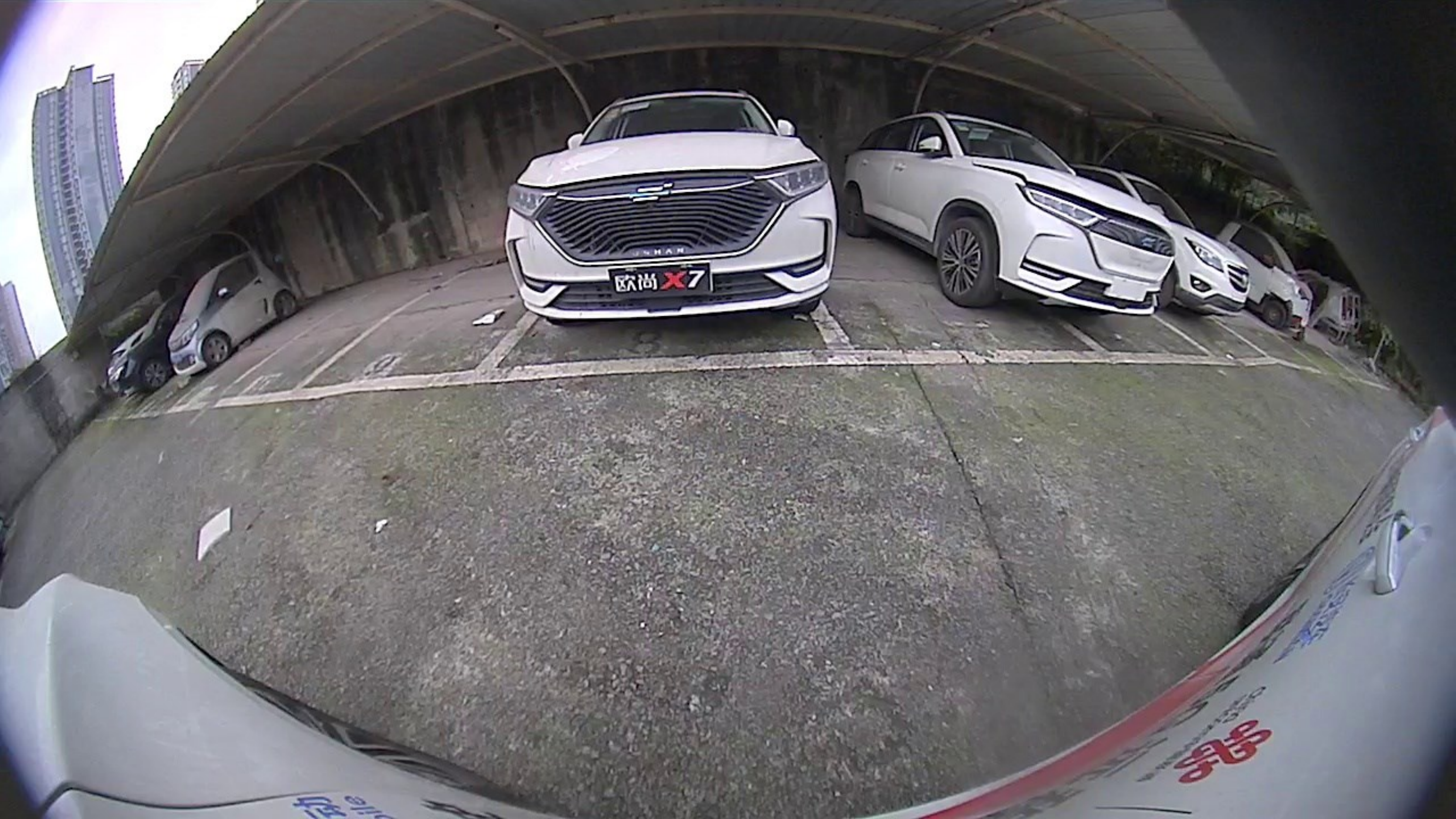}}\hspace{2mm}
   \subfigure[rear]{\includegraphics[width=4cm]{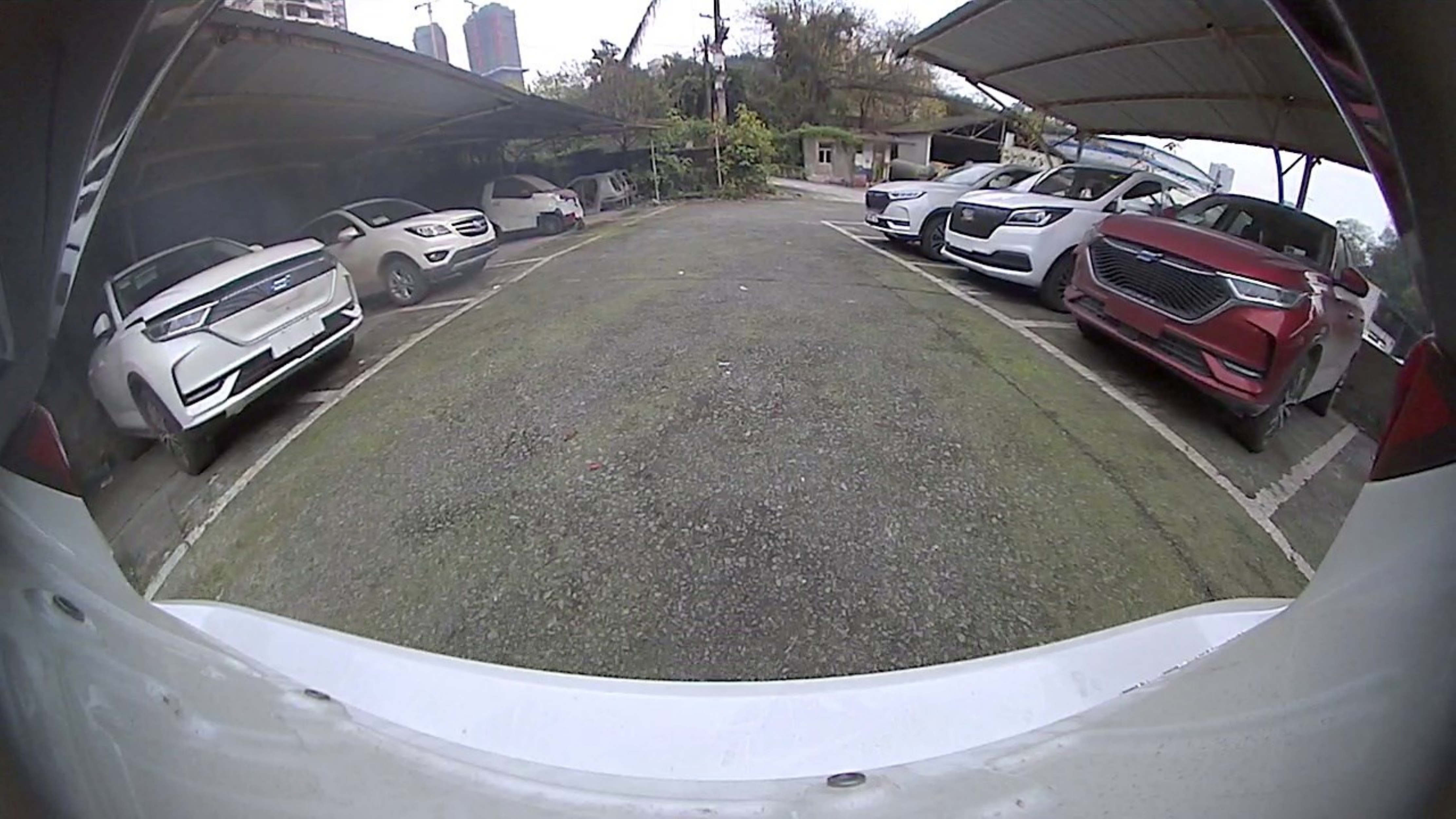}}

   \caption{(a) , (b) , (c) and (d) are the fisheye images captured by the front camera, left camera, right camera and rear camera while the vehicle is moving in the parking lot.}
   \label{fig:Figure09}
\end{figure}

      The WORD contains approximately 9,000 images, which were captured by a surround-view camera system, composed of four fisheye cameras. Examples are shown in Figure \ref{fig:Figure09}. There are approximately 2400 images from the front camera, 1800 images from the left camera, 2900 pictures from the right camera, and 2000 images from the back camera. The difference in the number of images from the different cameras is due to the fact that images that do not contain vehicles or do not have owner-member relationships are filtered out, which improves the quality of the WORD dataset. 
      
      The label of the WORD dataset consists of two parts. One part is the information of the edge boxes, containing seven attributes: x-coordinate and y-coordinate of the upper left corner of the edge box, x-coordinate and y-coordinate of the lower right corner of the edge box, score, class-ID, and box-ID; the other part is the owner-member relationship label, indicating which cars and wheels of the edge boxes have owner-member relationships. The visualisation of the owner-member relationship between the wheels and vehicles is shown in Figure \ref{fig:Figure04}.
 \end{appendices}

\end{document}